\documentclass[journal,twoside]{IEEEtran}

\usepackage{enumerate}
\usepackage{mathtools}
\usepackage{microtype}

\usepackage{pdflscape}

\usepackage{tabularx}
\usepackage{multirow}
\usepackage{textcomp}
\usepackage{csquotes}
\usepackage{booktabs}

\usepackage{xr-hyper}
\usepackage{hyperref}
\usepackage{amsmath}
\usepackage{amssymb}
\usepackage{lscape}
\usepackage{lineno}
\usepackage{array}
\usepackage{float}
\usepackage{todonotes}
\usepackage{tabu}
\usepackage{siunitx}
\usepackage{longtable}
\usepackage{pdflscape}
\usepackage{setspace}
\usepackage{textcomp}
\usepackage{rotating}
\usepackage{xcolor}
\usepackage{acronym}
\usepackage{multicol}
\usepackage{xtab}
\usepackage{multirow}
\usepackage{lipsum} 

\usepackage{natbib}
\usepackage{amssymb}
\usepackage{pifont}
\newcommand{\cmark}{\ding{51}}%
\newcommand{\xmark}{\text{\ding{55}}}

\DeclareUnicodeCharacter{2212}{-}

\usepackage{comment}
\def\BibTeX{{\rm B\kern-.05em{\sc i\kern-.025em b}\kern-.08em
    T\kern-.1667em\lower.7ex\hbox{E}\kern-.125emX}}

\ifCLASSINFOpdf
\else
 
\fi

\usepackage[a4paper, total={170mm,257mm}]{geometry}

\begin{document}

%


\title{Decoding ChatGPT: A Taxonomy of Existing Research, Current Challenges, and Possible Future Directions}


\author{\IEEEauthorblockN{Shahab Saquib Sohail\IEEEauthorrefmark{1},
Faiza Farhat\IEEEauthorrefmark{2}, 
Yassine Himeur\IEEEauthorrefmark{3}, 
Mohammad Nadeem\IEEEauthorrefmark{4},
Dag Øivind Madsen\IEEEauthorrefmark{5},
Yashbir Singh\IEEEauthorrefmark{6},
Shadi Atalla\IEEEauthorrefmark{3} and
Wathiq Mansoor\IEEEauthorrefmark{3}
}\\
\IEEEauthorblockA{\IEEEauthorrefmark{1}
Department of Computer Science and Engineering, School of Engineering Sciences and Technology, Jamia Hamdard, New Delhi 110062, India}\\
\IEEEauthorblockA{\IEEEauthorrefmark{2}Department of Zoology, Aligarh Muslim University Aligarh UP India}\\
\IEEEauthorblockA{\IEEEauthorrefmark{3}College of Engineering and Information Technology, University of Dubai, Dubai, UAE}\\
\IEEEauthorblockA{\IEEEauthorrefmark{4}Department of Computer Science, Aligarh Muslim University Aligarh UP India}\\
\IEEEauthorblockA{\IEEEauthorrefmark{5}University of South-Eastern Norway, Norway}\\
\IEEEauthorblockA{\IEEEauthorrefmark{6}Department of Radiology, Mayo Clinic, Rochester, MN, USA}\\
}


\maketitle

\begin{abstract}
Chat Generative Pre-trained Transformer (ChatGPT) has gained significant interest and attention since its launch in November 2022. It has shown impressive performance in various domains, including passing exams and creative writing. However, challenges and concerns related to biases and trust persist. In this work, we present a comprehensive review of over 100 Scopus-indexed publications on ChatGPT, aiming to provide a taxonomy of ChatGPT research and explore its applications. We critically analyze the existing literature, identifying common approaches employed in the studies. Additionally, we investigate diverse application areas where ChatGPT has found utility, such as healthcare, marketing and financial services, software engineering, academic and scientific writing, research and education, environmental science, and natural language processing. Through examining these applications, we gain valuable insights into the potential of ChatGPT in addressing real-world challenges. We also discuss crucial issues related to ChatGPT, including biases and trustworthiness, emphasizing the need for further research and development in these areas. Furthermore, we identify potential future directions for ChatGPT research, proposing solutions to current challenges and speculating on expected advancements. By fully leveraging the capabilities of ChatGPT, we can unlock its potential across various domains, leading to advancements in conversational AI and transformative impacts in society.
\end{abstract}


\IEEEpeerreviewmaketitle
\section{Introduction} \label{sec1}

In recent years, there has been a significant advancement in natural language processing (NLP) and artificial intelligence (AI) technologies, leading to the development of sophisticated language models capable of generating human-like text. Among these models, Generative Pre-trained Transformers (GPT) have gained tremendous attention and recognition for their ability to generate coherent and contextually relevant responses. GPT models have been successfully applied to various NLP tasks, including language translation, text summarization, and question answering \cite{guo2023can}. One prominent variant of the GPT model is Chat Generative Pre-trained Transformer (ChatGPT), a chatbot specifically designed to engage in conversational interactions with users \cite{gpt4,sohail2023using}. ChatGPT leverages the power of GPT to provide interactive and dynamic responses, mimicking human-like conversation. This innovative technology has opened up new possibilities in customer service, virtual assistants, and other applications where natural language understanding and generation are crucial \cite{cascella2023evaluating,deng2022benefits}.

As a result, there has been a growing interest in ChatGPT that extends far beyond the computer science discipline, with researchers from various backgrounds exploring its potential usefulness \cite{javaid2023chatgpt}. It has quickly gained worldwide attention and there is a lively discussion about its advantages and potential harmful effects. In a short span of time, ChatGPT has established itself as an excellent tool for accomplishing a variety of tasks, such as generating text on a given topic, obtaining information on a topic of interest, composing emails or messages with specific content and tone, modifying the structure or wording of a text, etc. \cite{salvagno2023can}. Additionally, it can generate code in multiple programming languages. Researchers have even used ChatGPT for scientific writing since it makes several parts of the academic writing process faster and more manageable, including article summarization, drafting, language translation, etc. \cite{lee2023can}.

\subsection{Architecture of ChatGPT}
ChatGPT, developed by OpenAI \cite{chatGPT}, is a language model that enables the creation of conversational AI systems capable of understanding and providing meaningful responses to human language inputs. Functioning as an AI-enabled chatbot, it employs algorithms to process user inputs and generate appropriate replies \cite{cao2023comprehensive}. ChatGPT has the ability to generate new responses or utilize pre-existing ones \cite{salvagno2023can}. To enhance its comprehension of user queries and generate accurate responses, ChatGPT undergoes continuous refinement using reinforcement methods, machine learning, and natural language processing techniques \cite{ouyang2022training}. In its own words (generated on March 29, 2023):

\textit{“I am ChatGPT, a language model developed by OpenAI, designed to generate human-like responses to a wide variety of questions and prompts. My purpose is to assist and interact with users in a conversational manner, providing helpful and informative responses to their inquiries.”}

To continually improve the reliability and accuracy of the model, ChatGPT incorporates reinforcement learning from human feedback (RLHF), allowing it to learn and understand human preferences through extended dialogues \cite{christiano2017deep,stiennon2020learning}. Additionally, researchers are actively exploring new technologies to enhance further its performance \cite{cao2023comprehensive,gpt4}. ChatGPT utilizes a transformer architecture consisting of encoder-decoder layers that collaborate to process and generate natural language text \cite{chen2023contextualized}. The architecture of ChatGPT comprises several vital components, such as the tokenizer, which divides raw text into smaller units called tokens for easier processing. The input embedding component then converts these tokens into high-dimensional vector representations \cite{wang2019language}.

\par The transformer architecture of ChatGPT consists of two main components: the encoder and the decoder \cite{budzianowski2019hello}. The encoder processes the input text hierarchically, creating representations at different levels of abstraction. On the other hand, the decoder generates the output text one token at a time, utilizing the input representations generated by the encoder.
An essential feature of the transformer architecture is the attention mechanism, which allows the model to selectively focus on different parts of the input text while generating the output \cite{jainknowledge}. This mechanism enhances the model's ability to capture relevant information and produce coherent output.
The output softmax layer is responsible for converting the high-dimensional vector representation of the output text into a probability distribution over the vocabulary of possible output tokens. This enables ChatGPT to generate high-quality and coherent natural language text.
The architecture of ChatGPT empowers it to excel in various tasks, including chatbots, language translation, and text completion, by generating accurate and meaningful responses.

Despite its popularity and usefulness, ChatGPT has raised concerns among researchers and practitioners due to its potential to generate content that, although seemingly reasonable, lacks factual accuracy \cite{borji2023categorical}. This issue can result in the production of counterfactual or meaningless responses, posing a serious threat to the reliability of online content. Additionally, the false narratives generated by ChatGPT can be easily mistaken as legitimate, especially by individuals who are unfamiliar with the topic at hand \cite{chatGPTFake}.
Researchers have been exploring and highlighting the potential harms associated with ChatGPT, including the propagation of stereotypes, biased responses, and dissemination of misleading information \cite{liang2021towards,nadeem2020stereoset}. Ethical concerns have also been raised regarding the use of ChatGPT, particularly when it is employed to create manipulated content that promotes misinformation and incites violence, potentially causing harm at both individual and organizational levels. Moreover, there are concerns that ChatGPT-generated content may infringe upon copyright and intellectual property rights \cite{deng2022benefits}. Furthermore, ethical considerations regarding the use of this tool for academic and scientific writing cannot be disregarded \cite{lee2023can}.

\subsection{Progress of ChatGPT research}
The initial iteration of ChatGPT referred to as GPT-1, was equipped with 117 million parameters and underwent training on a substantial corpus of text data \cite{ernst2022ai}. Subsequent versions, such as GPT-2, GPT-3, and the latest release, GPT-3.5, have experienced notable advancements by significantly augmenting the number of parameters. This augmentation has facilitated the generation of responses that are even more accurate and human-like.
An important breakthrough in ChatGPT is its capacity for zero-shot learning, which empowers the model to generate coherent responses to prompts it has never encountered before \cite{zhang2021commentary}. This remarkable capability is achieved through the utilization of unsupervised learning techniques and a novel training objective known as language modeling.

Despite the limitations of ChatGPT, its application has extended to various fields, including healthcare \cite{Abdel-Messih2023, sallam2023utility}, cyber security \cite{Mijwil202365}, environmental studies \cite{Rillig20233464}, scientific writing \cite{salvagno2023can, lee2023can}, education \cite{Tlili2023}, and others \cite{Wang2023575, Dowling2023, Biswas2023}. It is anticipated that the usage of ChatGPT will continue to grow in the future, with potential developments aimed at enhancing its capabilities \cite{Aljanabi202362, gpt4}. These developments may include real-time training of ChatGPT to improve its performance and the expansion of its domain-specific knowledge to make it more tailored and personalized for specific areas such as customer service, healthcare, business, or finance. Additionally, efforts can be made to address the issue of misinformation by ensuring that ChatGPT provides impartial and fair responses, thereby enhancing its trustworthiness and aligning with the growing importance of AI ethics and fairness considerations.

\subsection{Research Questions and Prime Contributions}
Writing a review about ChatGPT and its future contributions to recommender systems is crucial for synthesizing knowledge, identifying benefits and limitations, guiding future research, informing practitioners, and addressing ethical considerations. It serves as a valuable resource for advancing the field and maximizing the potential of ChatGPT in enhancing personalized recommendations. In doing so, this review article attempts to answer the following research questions (RQs):
\begin{itemize}
\item RQ1. What is the current state of ChatGPT research, including its architecture, advancements, and prime contributions?
\item RQ2. How diverse is the landscape of publications related to ChatGPT, and what are the recent trends in this research domain?
\item RQ3. What are the various applications of ChatGPT across different domains, such as healthcare, marketing and financial services, software engineering, academic and scientific writing, research and education, environmental science, and natural language processing?
\item RQ4. How can multimodal data (e.g., text, audio, visual) be leveraged to enhance the capabilities and performance of ChatGPT, and what are the key technical challenges in doing so?
\item RQ5. What are the main challenges, ethical considerations, potential risks, and ongoing research efforts in deploying GPT models in chatbot systems, and how are these being addressed to ensure fairness, transparency, explainability, and human-centered design?
\end{itemize}

It can be observed that many research areas around ChatGPT need to be explored. The current literature survey provides a detailed overview of research related to ChatGPT. There are existing review works on Large Language Models (LLM) \cite{melis2017state,chen2021evaluating} and AI Generated Content (AIGC) \cite{cao2023comprehensive} but they are very broad and do not take into consideration the specificities related to ChatGPT. To that end, this review work, a first of its kind, comprehensively performs a critical study of ChatGPT by covering 8 different applications, current issues and future challenges. The current study begins by reviewing the existing literature on ChatGPT and develops a taxonomy of areas/domains in which researchers have utilized this tool.

Additionally, the literature survey outlines the areas for improvement and presents potential challenges. Finally, the future applications of the tool and answers to its limitations are also explored. In summary, this survey makes several contributions, including:
\begin{itemize}
    \item A comprehensive review on ChatGPT.
    \item Presents and analyses the related literature.
    \item Highlights the areas in which this tool is predominately used.
    \item Concerns around ChatGPT and possible answers.
    \item Future enhancements and applications.
\end{itemize}

Moving forward, the proposed study has been compared with other surveys conducted on ChatGPT to identify the unique aspects and advancements it brings to the field. The analysis has considered various factors such as the applications covered, ChatGPT background, bibliometric analysis, research questions ChatGPT fine-tuning, open challenges and future research directions.
Table \ref{tab-comp} summarizes the output of this comparison.

\begin{table*}
\caption{Comparison of the proposed study with other surveys conducted on ChatGPT. We use tick marks (\cmark) to indicate the addressed fields and cross marks (\xmark) to denote missed fields.}
\label{tab-comp}
\scriptsize
\begin{tabular}{llllllllll}
\hline
Survey & Application & ChatGPT & Bibliomet- & RQs & ChatGPT & 
\multicolumn{3}{c}{Open challenges} & Future \\ \cline{7-9}
&  & background & ric analysis &  & fine-tuning & Intrinsic & Usage- & 
Ethical & Direction \\ 
&  &  &  &  &  &  & related & Concern &  \\ \hline
{\tiny \cite{lund2023chatting}} & Academia & \xmark & \xmark & \xmark & \xmark & %
\xmark & \xmark & \xmark & \xmark \\ 
{\tiny\cite{dwivedi2023so}} & Multidisciplinary & \cmark & \xmark & \cmark & \xmark
& \xmark & \xmark & \xmark & \xmark \\ 
{\tiny\cite{ray2023chatgpt}} & Multidisciplinary & \cmark & \xmark & \xmark & \xmark
& \cmark & \cmark & \cmark & \xmark \\ 
{\tiny\cite{kohnke2023chatgpt}} & Language & \xmark & \xmark & \xmark & \xmark & %
\xmark & \xmark & \xmark & \xmark \\ 
& teaching &  &  &  &  &  &  &  &  \\ 
{\tiny\cite{sifat2023chatgpt}} & Multidisciplinary & \cmark & \xmark & \xmark & %
\xmark & \xmark & \xmark & \xmark & \xmark \\ 
{\tiny\cite{rahman2023chatgpt}} & Academia & \cmark & \xmark & \xmark & \xmark & %
\xmark & \xmark & \cmark & \xmark \\ 
{\tiny\cite{temsah2023overview}} & Healthcare & \cmark & \xmark & \xmark & \xmark & %
\xmark & \xmark & \xmark & \xmark \\ 
{\tiny\cite{li2023chatgpt}} & Education & \cmark & \xmark & \cmark & \xmark & \xmark
& \xmark & \cmark & \xmark \\ 
{\tiny\cite{eggmann2023implications}} & Healthcare & \xmark & \xmark & \xmark & %
\xmark & \xmark & \xmark & \xmark & \xmark \\ 
{\tiny\cite{hill2023chat}} & Scientific publishing & \xmark & \xmark & \xmark & %
\xmark & \xmark & \xmark & \xmark & \xmark \\ 
{\tiny\cite{lo2023impact}} & Education & \xmark & \xmark & \xmark & \xmark & \xmark
& \xmark & \xmark & \xmark \\ 
{\tiny\cite{sallam2023chatgpt}} & Healthcare education & \xmark & \xmark & \xmark & %
\xmark & \xmark & \xmark & \xmark & \xmark \\ 
{\tiny\cite{gunawan2023exploring}} & Healthcare & \xmark & \xmark & \xmark & \xmark
& \xmark & \xmark & \xmark & \xmark \\ 
Ours & Multidisciplinary & \cmark & \cmark & \cmark & \cmark & \cmark & %
\cmark & \cmark & \cmark \\ \hline
\end{tabular}

\end{table*}

\par The rest of the paper is organized as follows. Section 2 highlights the variety of research domains that have published works on ChatGPT. Publication trends and taxonomy of ChatGPT literature are presented in Section 3. Section 4 discusses the applications of ChatGPT. Limitations of the tool and future enhancements are outlined in Section 5 and Section 6, respectively. Finally, concluding remarks are presented in Section 7.

\section{Survey Methodology} \label{sec2}
To conduct our literature review, we adopted the methodology proposed in \cite{kitchenham2004procedures}. It is crucial to recognize that identifying the need for a review is just as important as the review itself. The rapid dissemination of ChatGPT research resulted in a diverse research landscape due to its wide publicity and acceptance at different levels in our daily lives. Our study highlights the need for a comprehensive review that outlines the various aspects of its usage for different applications, its limitations and potential future directions.  \newline
After running search queries, inclusion (IC) and exclusion (EC) criteria were created to filter out the retrieved articles from the Scopus database. Material obtained may include multiple papers that match the search query but are irrelevant to our study. We have also performed a screening of the articles by going through the abstract and assessing its relevance to the title under consideration. The Inclusion and exclusion criteria comprise the following:
\newline
Inclusion criteria:
\begin{itemize}
    \item IC-1: English must be the medium of the paper.
    \item IC-2: The foundation of the paper must be a peer-reviewed publication, such as one from a workshop, journal, book, conference, etc.
    \item IC-3: Articles that contain keyword “chatgpt” or “chat-gpt” in their abstract or title.
    \item IC-4: Articles that discuss ChatGPT.
    \item IC-5: Articles that are published till March 25, 2023.
\end{itemize}

Exclusion criteria:
\begin{itemize}
    \item EC-1: Short papers.
    \item EC-2: Articles that did not contain keyword “chatgpt” or “chat-gpt” in their abstract or title.
    \item EC-3: Articles discussing only GPT or generative AI and not exclusively chat GPT.
    \item EC-4: Duplicate articles.
    \item EC-5: Earlier versions with errata.
\end{itemize}

\begin{figure*}[t!]
\centering
\includegraphics[width=0.9\linewidth]{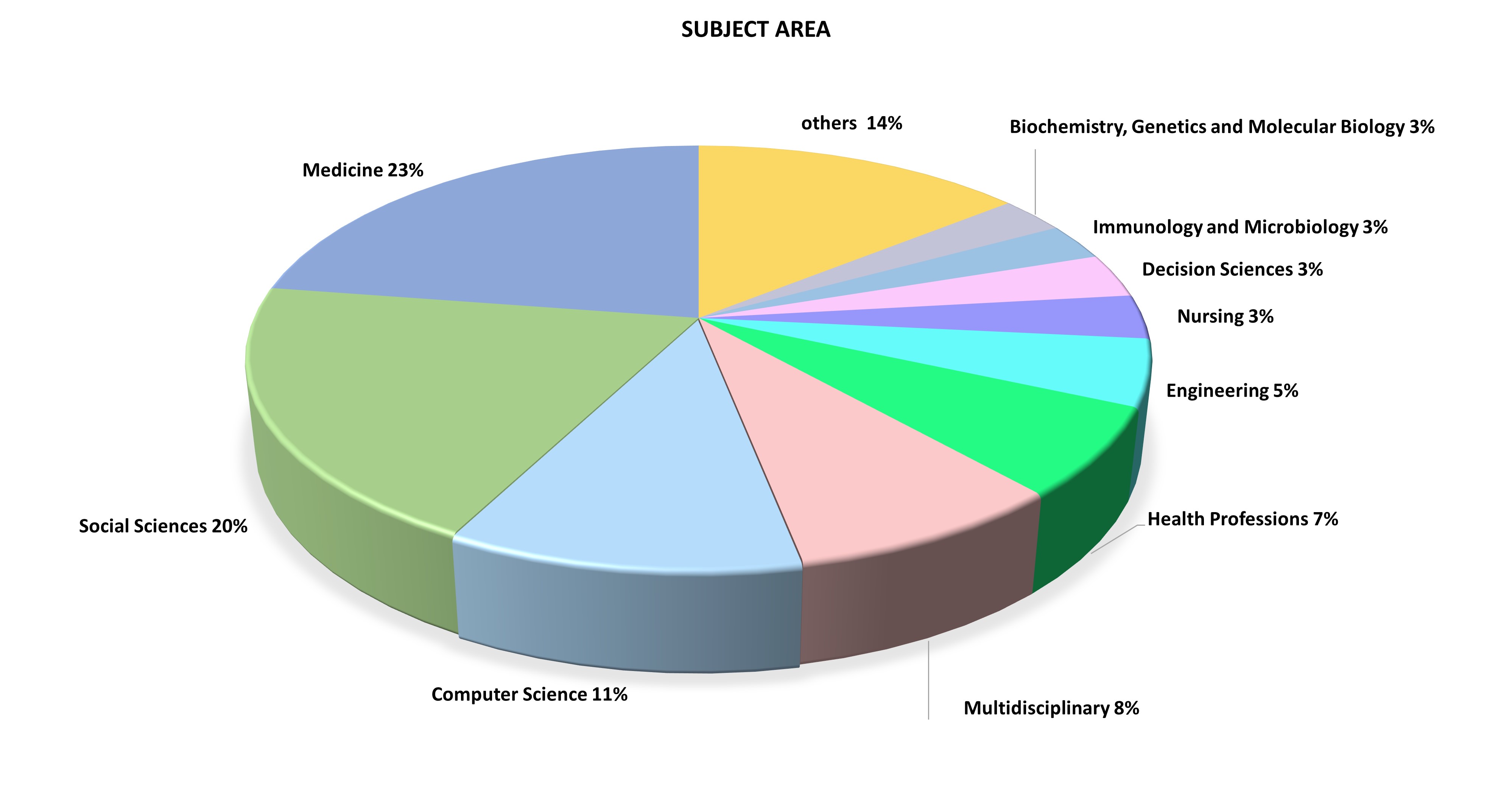}
\caption{Publications on ChatGPT on the basis of subject area}
\label{fig:subject}
\end{figure*}

\begin{figure*}[t!]
\centering
\includegraphics[width=1.0\linewidth]{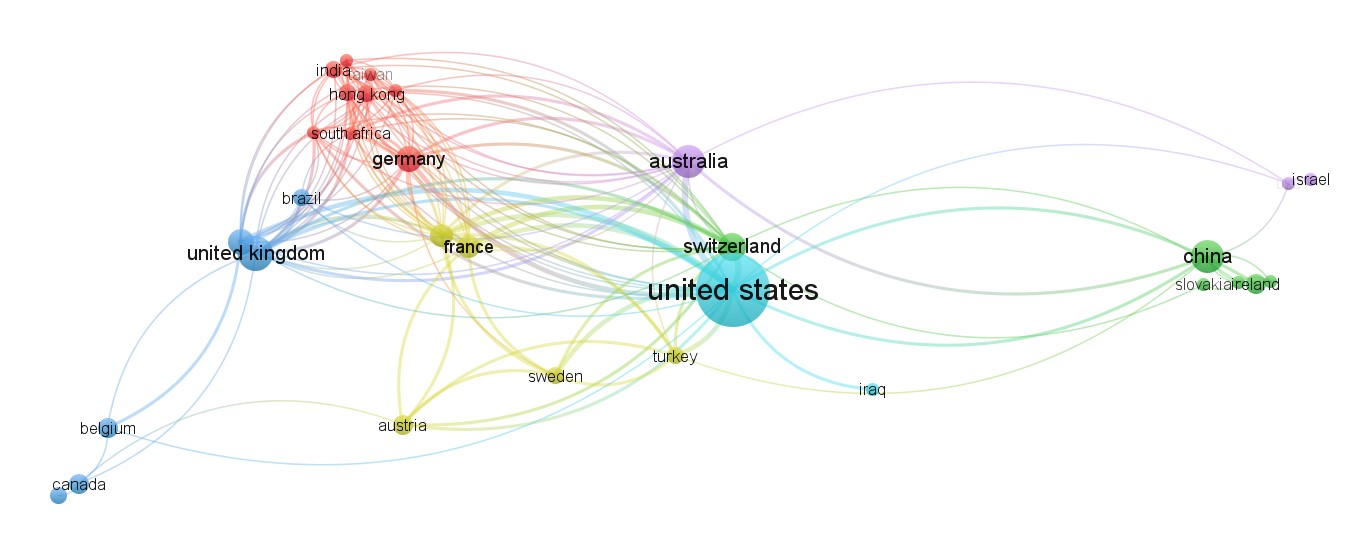}
\caption{Top contributing countries}
\label{fig:countries}
\end{figure*}

After applying the specified criteria, a total of 109 articles were included in the analysis. These articles involved contributions from 349 authors representing 53 different nations, indicating a wide range of international participation in the literature on ChatGPT. The publications were further examined and categorized based on their respective subject areas. Notably, the field of medicine had the highest representation, accounting for 23\% of the total publications. This was followed by social sciences (20\%) and computer science (11\%) (see Figure \ref{fig:subject}). Multidisciplinary subjects and health professionals made up 8\% and 7\% of the total corpus, respectively.
\vskip2mm
Figure \ref{fig:countries} provides a visual representation of the collaborative network among countries contributing to the topic of ChatGPT. The size of each circle corresponds to the number of documents produced by that country, while the connecting lines represent collaboration links between countries. The thickness of the connecting lines corresponds to the frequency of collaboration between the respective countries. The United States ranked first as the country of origin for published articles, with a total of 33 publications in the corpus. It was followed by the United Kingdom with 10 publications, and Australia and China with 9 publications each. In terms of collaborations, the United States had the most extensive network, collaborating with 24 different countries, accounting for over 18\% of the total corpus. Switzerland ranked second in terms of collaboration, with 20 collaborative countries, followed by Australia with 19 collaborations, and the United Kingdom with 18 collaborations (see Figure \ref{fig:countries}).

\section{Diversity of publication on ChatGPT}

Following its launch, ChatGPT quickly gained a lot of popularity in a variety of disciplines \cite{Cox2023, dwivedi2023so,Tlili2023}, including academia \cite{Chen2023} and science \cite{Morreel2023}. It is being used for a wide range of applications from content creation, translation, writing essays, and computer coding to research assistance. Research scholars have been using ChatGPT in a variety of ways, including writing scientific literature, gathering data, drafting abstracts for research papers \cite{Else2023423}, and looking for medical diagnostic advice \cite{Hirosawa2023}. Even entire scientific articles have occasionally been generated using it, with authorship provided. However, there has been criticism and backlash from many people as a result of ChatGPT's instances of generating inaccurate information or being perceived as a threat to the integrity of plagiarism-free scientific texts.

As far as publication avenues (journals and conferences) are concerned, Nature Journal tops the list with 13 articles. Next, Accountability in Research has published 4 articles. Other journals with multiple publications include JMIR Medical Education (3), Journal of Educational Evaluation for Health Professions (3), and The Lancet Digital Health (3). Iraqi Journal for Computer Science and Mathematics has also published 2 documents as well. However, no articles have been published in top conferences on NLP, possibly because it usually takes more time for conference announcement and acceptance, and consequently gets published online.

\section{Recent trend of publications related to ChatGPT} \label{sec2}
A thorough literature review was conducted, and 109 articles were found after searching the Scopus database for relevant articles on ChatGPT. Following the classification of the retrieved articles using either their abstracts or full texts, it was discovered that there were mainly three categories of articles published up until March 25th, 2023: 1) evaluations of ChatGPT, 2) predictions made using ChatGPT, and 3) reviews on ChatGPT. The biggest cohort consisted of ChatGPT assessments across various domains. A total of 68 articles were published to evaluate ChatGPT's ability to gauge its proficiency in providing accurate answers or the depth of its knowledge. Making predictions using ChatGPT for different fields is the subject of the second-largest group of articles (39) and the fewest reviews (10 publications).

\begin{figure*}[t!]
\centering
\includegraphics[width=1.0\linewidth]{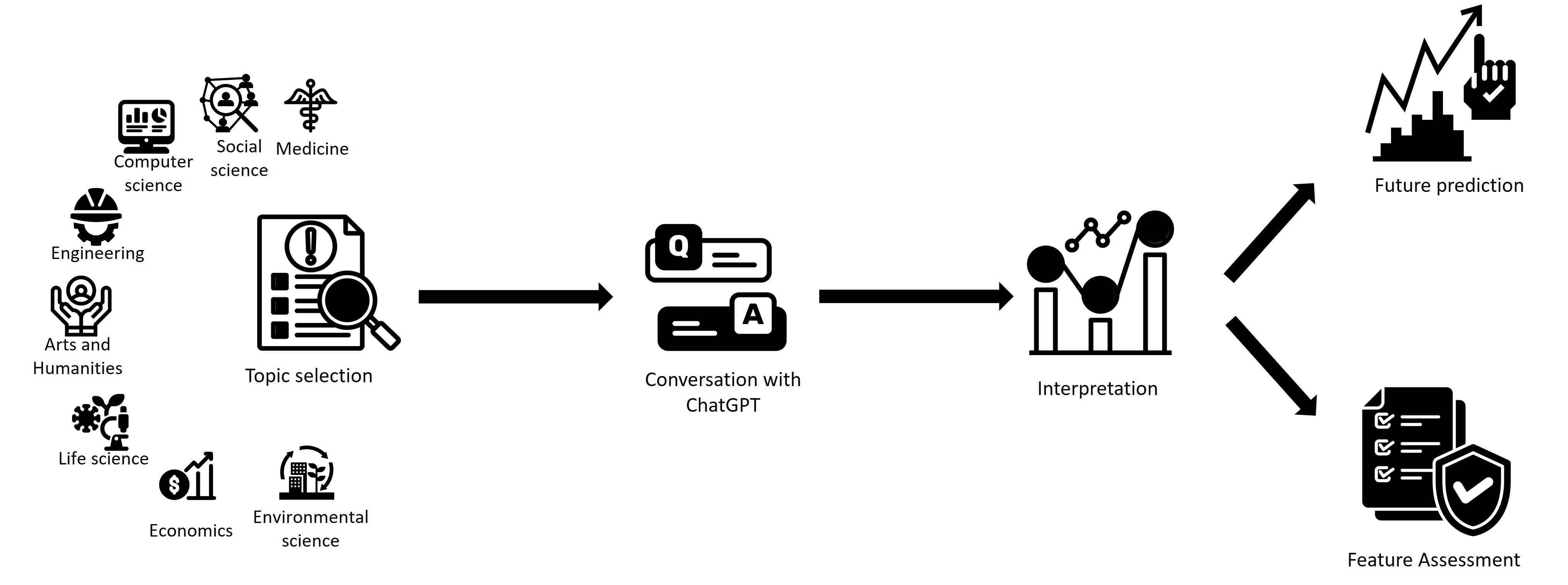}
\caption{A common trend identified in the reported chatGPT research}
\label{fig:commonTrend}
\end{figure*}

\textcolor{black}{Upon reviewing the entire body of literature related to ChatGPT, a prevalent pattern has emerged in the structure of most articles, as shown in Figure \ref{fig:commonTrend}. Typically, authors select a topic of interest, ask ChatGPT questions, and then provide interpretations based on its responses. These interpretations can be broadly categorized into two types: those that predict the future of the topic in the context of ChatGPT and those that assess ChatGPT's features and their potential impacts.}

Furthermore, it has been noted that the majority of publications center around either ChatGPT's feature assessment as a scientific writing assistant or predictions about the future of the education system. As a result, many authors have expressed concerns regarding the ethical and scientific implications of ChatGPT's features and their potential negative effects on scientific academia. Meanwhile, several articles have explored how ChatGPT could impact learners and educators, as well as the ethical considerations that must be taken into account when using it.

\begin{figure*}[t!]
\centering
\includegraphics[width=0.9\linewidth]{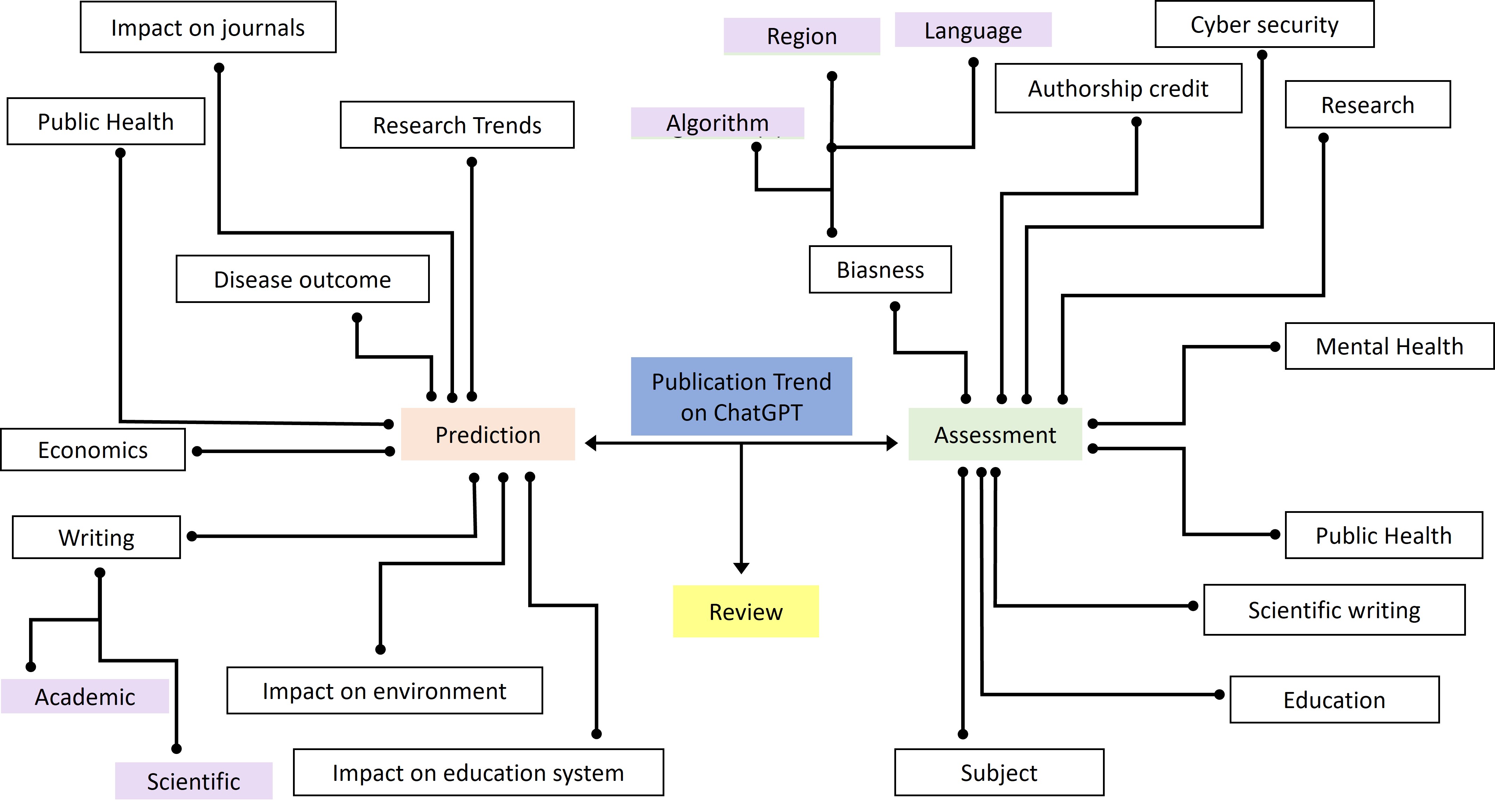}
\caption{Taxonomy of literature on ChatGPT}
\label{fig:TaxonomyChatGPT}
\end{figure*}

The different assessments included those for bias \cite{Wang202334,Wang2023575}, for study \cite{Cooper2023}, for mental health \cite{Prada2023532}, for public health \cite{Jungwirth2023}; \cite{Biswas2023}, etc. It is noted that the majority of authors have evaluated ChatGPT for its capacity for scientific writing (30 publications), whether in terms of producing data \cite{maddigan2023chat2vis}, writing a scientific article \cite{D'Amico2023663,Anderson2023, farhat2023trustworthy}, or fetching references for particular subjects \cite{gravel2023learning,alkaissi2023artificial, farhat2023trustworthy}. The potential of ChatGPT to make a contribution to the field of education from the perspectives of learners \cite{Gašević2023}, educators \cite{Lim2023}, and mentors \cite{Naumova2023,Johinke2023,Rospigliosi20231} was the second biggest issue. The third biggest group of researchers from various fields questioned ChatGPT to evaluate its research ability to contribute to the field of research. Additionally, some authors have evaluated it for subject expertise by posing it with various test questions, such as those from bar exams \cite{bommarito2022gpt}, medical exams \cite{Gilson2023}, etc. Some have also asked questions based on their extensive understanding of various fields, including parasitology \cite{Huh20231,Šlapeta2023}, clinical diagnosis \cite{Hirosawa2023}, environmental sciences \cite{Rillig20233464}, public health \cite{Biswas2023}, etc. A few authors also questioned ChatGPT’s efficiency in cyber security \cite{Mijwil202365} and its potential for biases due to algorithms \cite{Wang202334}, region \cite{Wang2023575}, and language \cite{Seghier2023216}. Some authors have also used ChatGPT for writing research papers and credited it as an author \cite{Mijwil202365,Aljanabi202362}, while others have questioned the validity of such authorship \cite{Stokel-Walker2022,Siegerink2023} (Figure \ref{fig:TaxonomyChatGPT}).

The second most popular research publication subject was prediction through ChatGPT. Numerous researchers have forecasted how ChatGPT will affect various areas, including the educational system \cite{Choi2023,Lim2023} (12 publications). In terms of predictions, disease diagnosis \cite{Khan2023605,Mann2023221}; \cite{DiGiorgio2023} came in second. Clinicians and medical professionals probed ChatGPT with hypothetical or real-world patient symptoms to determine whether or not it would aid in the diagnosis or treating patients. It was also very common among researchers to predict research trends in specific research fields using ChatGPT \cite{Tong2023220}, which would be helpful for aspiring researchers to pursue their research. Concern has also been expressed by a group of scholars who anticipate how ChatGPT could impact writing in both academic \cite{Perkins2023} and scientific fields \cite{Ali2023,Tregoning2023}. After conducting an extensive literature survey, it was discovered that only one out of the ten reviewed articles published so far discusses AI-Driven Conversational Chatbots, including ChatGPT, from 1999-2022 \cite{Lin2023}. However, it should be noted that this article does not solely focus on ChatGPT. Therefore, it can be inferred that a systematic literature review that is exclusively dedicated to ChatGPT has not yet been published. It is worth mentioning that the remaining nine reviewed articles focused on the efficiency of ChatGPT. This was either accomplished through direct assessment or prediction using ChatGPT \cite{Budler2023,Thurzo2023,Haman2023}.

\textcolor{black}{It is also important to note that Prompt Engineering is turning out to be an emerging dimension related to ChatGPT \cite{white2023prompta}. Researchers explored various techniques for designing effective prompts to elicit desired responses from the tool. This involved providing explicit instructions or giving example outputs to guide the tool's behavior \cite{,white2023chatgptb}. Another important trend was the development of diversity-promoting approaches. To address issues related to generating repetitive or generic responses, researchers experimented with techniques to encourage the generation of diverse and creative outputs \cite{cao2023comprehensive}. Ethical considerations are discussed in the majority of the papers either in detail or briefly. As AI systems interact with users, concerns about bias, fairness, and ethical issues have become prominent. Studies focused on developing methodologies to mitigate bias in language models and ensure they adhere to ethical guidelines, such as promoting fairness and avoiding harmful or offensive responses \cite{ray2023chatgpt}. It is obvious that the ethical dimension of ChatGPT will remain a hot topic in the future.}

\section{Applications of ChatGPT}
This section provides an overview of the primary applications of ChatGPT and generative chatbots in general. Furthermore, Table \ref{tab2} offers a summary of the existing research on ChatGPT, outlining the area of study, applications, objectives, and key findings of each study. \textcolor{black}{One of the common applications of ChatGPT is as a personal assistant \cite{bakker2022fine}. In different domains, the tool is customized to serve the needs of specific domains. For example, in the domain of healthcare, ChatGPT is used as a virtual medical assistant to provide patients with information about symptoms, medications, and general healthcare advice. They can help triage patients by asking relevant questions and suggesting appropriate next steps \cite{Perkins2023,wang2023chatgpt}. In marketing, ChatGPT works as a conversational chatbot to handle customer inquiries, provide product recommendations, and assist with order tracking. This helps streamline customer support processes, reduces wait times, and improves customer satisfaction \cite{cascella2023evaluating}. ChatGPT is turning out to be a Code Assistant and Debugger. It helps software developers by providing code suggestions, debugging assistance, and answering programming-related questions \cite{aljanabi2023chatgpt}. This improves development productivity and facilitates knowledge sharing among developers.}

\subsection{Healthcare}
Although ChatGPT has access to limited medical data, it has demonstrated performance in medical licensing exams equivalent to that of an undergraduate third-year medical student. As a result, there have been urgent discussions within the medical field about the impact of ChatGPT. Stokel-Walker and van Noorden describe in their article the implications of generative AI for science and how ChatGPT can answer some open-ended medical queries nearly as well as an average human physician \cite{sohail2023chatgpt} but with some shortcomings and unreliabilities \cite{stokel2023chatgpt}.
Clinicians often rely on complex information to make decisions, but in telemedicine, the available information is typically limited to language only, making it a potential candidate for interventions using LLM \cite{roosan2016big}. However, ChatGPT, a popular LLM, has limitations as it cannot ask questions to clarify questions or scenarios. This becomes particularly challenging in infection consultation, which requires the integration of clinical information with knowledge related to antimicrobial resistance and microbial ecology. To explore this further, the researchers asked ChatGPT for antimicrobial advice in eight hypothetical infections scenario-based questions and evaluated the appropriateness, consistency, safety, and antimicrobial stewardship implications of its responses. Based on this evaluation, they constructed an LLM medical safety assessment framework to assess the safety of LLM responses \cite{bashshur2020telemedicine,howard2023chatgpt}

With a few notable exceptions, ChatGPT demonstrated appropriate recognition of natural language \cite{howard2023chatgpt}. The model's understanding of scenarios was evident from the accurate summaries provided at the beginning of its responses. However, important situational aspects were not always well distinguished from unimportant ones. ChatGPT was able to recognize clinically important factors when they were explicitly provided, but as the complexity of the scenarios increased, it missed relevant issues \cite{howard2023chatgpt}.
ChatGPT's responses were coherent, and their spelling and grammar were appropriate. The model's answers included a summary of its understanding of the scenario and question, management options, and disclaimers that reflected its information sources, which were similar to the format of patient information websites. ChatGPT often repeated questions verbatim, including any errors, although it occasionally noticed and corrected them. The information provided by ChatGPT was consistent, and it did not repeat the same advice in a single response. However, the advice provided sometimes changed when questioned repeatedly.

The study in \cite{howard2023chatgpt} evaluated ChatGPT's ability to provide antimicrobial advice and found that while its antimicrobial spectra and regimens were appropriate, duration appropriateness varied and source control was often disregarded. ChatGPT also had deficits in situational awareness, inference, and consistency and sometimes gave dangerous advice despite prompting. However, the study proposes a modifiable qualitative framework to address these issues and urges clinicians to familiarize themselves with this new technology. The author declares several competing interests, while all other authors have none.

\subsection{Marketing and financial services}
The use of AI in banking has become increasingly important in recent years, with ChatGPT offering opportunities for back-end operations, data analysis, and personalized customer offers. AI can be used to understand consumer needs and create effective marketing strategies, but there are limitations to relying solely on ChatGPT due to high regulations in the financial services sector. Human involvement is necessary to verify the trustworthiness of insights and offers, and banks must invest in infrastructure and human resources to integrate AI into their digital transformation strategies.

Many studies, such as \cite{dwivedi2023so,geerling2023chatgpt,street2023let,rathore2023future}, have already investigated the potential of using ChatGPT in banking operations, both for back-end data analysis and marketing communication strategies, as well as for front-end operations to engage with customers directly. While there have benefits to using ChatGPT, there are concerns about trust and its impact on customer well-being. Banks will need to invest in training staff, educating customers about the technology, and knowing when and how far to push it. Trust in service provision will be crucial, and there are implications for convenience, promptness, and accurate decisions.

\clearpage

\onecolumn 

\begin{longtable}{
    m{2cm}
    m{2cm}
    m{3.5cm}
    m{4cm}
    m{3cm}
}
\caption{Summary of the reported work on ChatGPT describing area of the study, applications, objectives and key findings of the research.}
\label{tab2}\\
\hline
Area of study & Application & Objectives & Key findings & References \\
\endfirsthead
\hline
\multicolumn{5}{c}{{Table \thetable\ (Continue)}} \\
\hline
Area of study & Application & Objectives & Key findings & References \\
\endhead
\hline
\endfoot

\multirow{7}{*}{Medicine} & Scientific writing & Evaluate the potential for medical study fabrication using AI-generated ChatGPT & The Combination of ease of creating fabricated work, the challenging detection of fraudulent publications, and absence of AI-based detection technologies creates an environment that facilitates fraudulent research. Researchers and practitioners can effectively utilize ChatGPT technology while avoiding any unintended consequences by developing a comprehensive understanding of its capabilities and limitations. & \cite{kitamura2023chatgpt,Elali2023,Liebrenz2023e105,biswas2023chatgpt,Arif2023,marchandot2023chatgpt,arif2023future,ufuk2023role,Cascella2023,lubowitz2023chatgpt} \\
\cline{2-5}
& Mental health care & Investigate the opportunities and challenges of ChatGPT in mental health care. Evaluate emotion-enhanced prompting and ChatGPT for mental health analysis. Evaluate ChatGPT for NLP-based mental health applications. & ChatGPT can offer emotional support and engagement to individuals with mental health concerns. ChatGPT can help assess the risk level of individuals experiencing mental health crises by analyzing conversations. ChatGPT can offer emotional support and engagement to individuals with mental health concerns. & \cite{singh2023artificial,yang2023evaluations,lamichhane2023evaluation,bhattacharyya2023chatgpt,van2023artificial,aminah2023considering,qiu2023smile,okan2023ai,uludag2023testing} \\
\cline{2-5}
& Education \& Examination & Evaluate the performance of ChatGPT in Medical Physiology Examination Phase I MBBS & Prior to implementation, the accuracy, source, and reliability of the information should be validated by expert faculty and clinicians in order to rely on ChatGPT for education or medical practices & \cite{subramani2023evaluating,sedaghat2023early,hisan2023chatgpt,sallam2023chatgpt,hisan2023chatgpt,fatani2023chatgpt,lee2023rise,alser2023concerns,talan2023role,sallam2023utility,Khan2023605,kung2023performance,gilson2023does} \\
\cline{2-5}
& Cardiology and Vascular Pathologies & Evaluate the precision of ChatGPT to the Basic Life Support (BLS) and Advanced Cardiovascular Life Support (ACLS) examinations. Explore ChatGPT for information of cardiopulmonary resuscitation. & ChatGPT did not reach the passing threshold for any of the exams. ChatGPT can assist researchers in analyzing large datasets related to cardiovascular diseases. ChatGPT can serve as a tool for healthcare providers by offering decision support in managing cardiovascular diseases. ChatGPT can help analyze patient data, such as medical records, family history, lifestyle factors, and biomarkers, to assess an individual's risk for cardiovascular diseases. & \cite{Fijačko2023,moons2023chatgpt,harskamp2023performance,skalidis2023chatgpt,nakaya2023chatgpt,harskamp2023performance,haver2023appropriateness,van2023response,fijavcko2023can,biswas2023reducing,ahn2023exploring,williams2023will} \\
\cline{2-5}
& Medical Licensing Examination & Assess the performance of ChatGPT in Medical Licensing Exams across multiple countries. & ChatGPT achieves greater than 60\%, passing score. ChatGPT demonstrates its versatility as a medical assistant by effectively analyzing real-world medical issues in a manner that is accessible, user-friendly, and adaptable. ChatGPT shows potential to support clinical decision-making in Japanese healthcare settings, but caution is needed due to performance improvements required. ChatGPT's knowledge and interpretation in the Chinese Chinese National Medical Licensing Examination (NMLE) are below medical students' level, but deep learning may enhance its abilities. & \cite{Perkins2023,wang2023chatgpt,kasai2023evaluating,gilson2022well,kaneda2023can,wu2023qualifying,bhayana2023performance} \\
\cline{2-5}
& Clinical Diagnosis & Assess the reliability of ChatGPT generated clinical scenario differential-diagnosis lists & The total rate of the correct diagnoses within ten differential-diagnosis lists generated by ChatGPT was more than 90\%. & \cite{Hirosawa2023} \\
\cline{2-5}
& Gastroenterology research & Assess the potential of ChatGPT for GI research & ChatGPT has the potential to contribute to the advancement of gastroenterology by generating high-quality research questions & \cite{Lahat20234164} \\
\hline
& Nursing & Investigate the potential application of ChatGPT in the field of nursing and caregiving services. & ChatGPT can provide valuable insights into the future of nursing. & \cite{gunawan2023exploring,alkhaqani2023chatgpt,odom2023role,moons2023chatgpt} \\
\hline
Business & Multidisciplinary & Evaluate ChatGPT in the context of education, business, and society & Enacting new laws to regulate these tools is crucial & \cite{dwivedi2023so,george2023review,chuma2023business,beerbaum2023generative} \\
\hline
Business \& Management & Management \& Education & Examine the use of ChatGPT in management education & In the future of education, generative AI should be welcomed rather than avoided & \cite{Lim2023} \\
\hline
Business and Economics & Scientific Research & To assess the use of ChatGPT for the research & ChatGPT can produce plausible-appearing research papers for reputable journals & \cite{Dowling2023,kshetri2023chatgpt,mcgee2023top,mcgee2023capitalism,george2023review} \\
\hline
\multirow{2}{*}{Life Science} & Parasitology test & Evaluate ChatGPT's understanding and comprehension skills of Korean medical students for parasitology test & The knowledge and ability to analyze results of parasitology test by ChatGPT were not yet on par with those of Korean medical students & \cite{Huh20235,huh2023chatgpt,vslapeta2023chatgpt} \\
\cline{2-5}
& Synthetic Biology & Check accuracy of the ChatGPT generated information before spreading. Investigate how cutting-edge AI can help practitioners of synthetic biology & ChatGPT has demonstrated its potential in various aspects of medicine, including supporting translational medicine, drug development, medical reporting, diagnostics, and treatment plans. Ethically adhering to the use of ChatGPT and other large language models (LLMs), computational biologists can enhance their efficiency, leading to accelerated scientific discovery in the field of life sciences. & \cite{Tong2023220,agathokleous2023use,lubiana2023ten,cahan2023conversation} \\
\hline
Chemistry & Global epidemiology & Interpret epidemiological relation between particulate matter and mortality risks & Prolonged questioning could be beneficial in enhancing and improving both the kinds of human reasoning and argumentation imitated by current LLMs. & \cite{Cox202399} \\
\hline
\multirow{3}{*}{Education} & Teaching \& Learning & Examine the use of ChatGPT in education. Understand the potential benefits of ChatGPT in promoting teaching and learning. & More guidelines on how to use ChatGPT safely in education should be created, and it should be used with more caution. & \cite{Tlili2023,alafnan2023chatgpt,lo2023impact,tlili2023if,mhlanga2023open,qadir2023engineering,firat2023chat,kasneci2023chatgpt,tlili2023if,qadir2023engineering} \\
\cline{2-5}
& Education system and library science & Assess ChatGPT potential impact on academia and libraries & Use ChatGPT responsibly and ethically to create new knowledge and educate future professionals & \cite{Lund2023} \\
\hline
\multirow{3}{*}{Language} & Learning Foreign Languages & Explore the development of chatbot systems and the principal approaches and data sets employed in their creation. & NLP technologies are being used to create conversational chatbots to mimic the conversational proficiency of humans. & \cite{Lin2023,hong2023impact,kohnke2023chatgpt,kasneci2023chatgpt,ali2023impact,lai2023chatgpt} \\
\cline{2-5}
& Vocabulary Expansion & Investigate the use of ChatGPT in (i) word meaning clarification, (ii) synonyms and antonyms, (iii) contextual usage, (iv) collocations and word associations, (v) idioms and figurative language, (vi) specialized vocabulary, and (vii) Word usage tips. & ChatGPT can assist learners in expanding their vocabulary. Learners can inquire about word meanings, synonyms, antonyms, and usage examples, allowing them to acquire new words and enhance their lexical knowledge. & \cite{huang2023role} \\
\cline{2-5}
& Linguistic Ambiguity Analysis & Study the ChatGPT strengths and weaknesses for linguistic ambiguity analysis & ChatGPT helps identify instances of linguistic ambiguity in text or speech. It can recognize when a word, phrase, or sentence has multiple possible interpretations, leading to potential confusion or miscommunication. & \cite{huang2023chatgpt,ortega2023linguistic,huang2023chatgpt,gao2023human} \\
\hline
Academia & Academic Writing & Investigating the use of ChatGPT in (i) enhanced writing productivity, language refinement and fluency, (iii) Knowledge synthesis and content generation, and (iv) revision and editing support. & While ChatGPT provides improved and faster academic writing, it also introduces several challenges, such as (i) overreliance on AI suggestions, lack of context awareness, and ethical considerations. Further research is required to address limitations and challenges, studying their impact on students' writing skills, perceptions, and academic performance. Context-aware AI models aligned with academic conventions should be developed. & \cite{dergaa2023human,alkaissi2023artificial,cotton2023chatting,bom2023exploring,chen2023chatgpt,alafnan2023chatgpt,donmez2023conducting,tomlinson2023chatgpt,lund2023chatgpt,aczel2023transparency,frye2022should,chen2023chatgpt} \\
\hline
Robotics & Robotic Process Automation & Explore the potential of generative AI, especially ChatGPT in robotics. Investigate the ethics of using ChatGPT for robotic process automation. Investigate the prospective role of Chat GPT in the military. Study the use of ChatGPT-empowered long-step robot control in various environments. Explore the level of trust in human-robot collaboration utilizing ChatGPT. & ChatGPT or similar AI models can assist in task execution and control of robots. ChatGPT can aid in diagnosing issues and providing troubleshooting guidance for robotic systems. ChatGPT can help establish shared understandings, allocate tasks, and assist in real-time communication, thereby improving the overall efficiency and performance of robotic teams. ChatGPT can act as a user assistance tool, providing guidance, explanations, and educational resources to users interacting with robots. & \cite{vemprala2023chatgpt,wake2023chatgpt,biswas2023prospective,beerbaum2023generative,you2023robot} \\
\hline
Environment & Global Warming & Explore the impact of ChatGPT on global warming in terms of (i) environmental data analysis and interpretation, (ii) environmental education and awareness, (iii) environmental policy and planning, (iv) climate change modeling and projections, (v) natural resource management, and (vi) Environmental monitoring and early warning systems. & Emphasize the significant role that AI and natural language processing technologies, represented by ChatGPT, can play in advancing our understanding of climate change and improving the accuracy of climate projections. ChatGPT facilitates informed decision-making, enhances environmental awareness, and aids in the development of sustainable practices for a more resilient and ecologically balanced future. & \cite{biswas2023potential} \\
\hline
Smart Vehicles & & Explore the potential impact of ChatGPT for intelligent vehicle research. Explore the conversation with ChatGPT on interactive engines for intelligent driving. & Use ChatGPT's information can be updated and corrected, but it may not always reflect the latest knowledge. & \cite{Gao20231,du2023chat,zhang2023hivegpt,chen2023feedback,lei2023chatgpt,zheng2023chatgpt,wang2023linguistic} \\
\hline
\end{longtable}

\clearpage
\newpage

\twocolumn 
\subsection{Software Engineering}
Software engineering is a broad field consisting of sub-processes such as software development, designing, testing, coding, etc. ChatGPT has been shown to assist in all these sub-domains of software engineering \cite{sobania2023analysis,aljanabi2023chatgpt,surameery2023use, white2023prompta,white2023chatgptb}. ChatGPT has a significant benefit in coding, as it can process human inputs, allowing software developers to supply code snippets or commands in a conversational way rather than providing specific keywords or phrases. This feature can enhance the coding experience for less-skilled programmers, making it more user-friendly and intuitive \cite{aljanabi2023chatgpt}. Researchers have used ChatGPT for automated fixing of programming of software bugs \cite{sobania2023analysis,surameery2023use}. They evaluated the ChatGPT's bug-fixing ability on the standard QuixBugs benchmark set and compared its performance with existing methods. They concluded that ChatGPT's bug-fixing performance is comparable to standard deep learning techniques/tools such as CoCoNut and Codex and superior to standard approaches. Since it is a conversational tool, its bug-fixing ability was improved further by providing it hints.

\par Managing the architecture of software-intensive systems can be a difficult and intricate process that involves integrating various perspectives from designers, stakeholders, automation tools, and other factors to create a roadmap for guiding software development and assessment. \cite{ahmad2023towards} used ChatGPT for analyzing, synthesizing, and evaluating the architecture of a services-oriented software application. They concluded that in the presence of human observation, ChatGPT could be used in place of a full-time human architect to carry out the process of architecture-centric software engineering. Furthermore, \cite{white2023prompta,white2023chatgptb} proposed a prompt engineering framework using ChatGPT for automating the process of software development, including the creation of API specifications, decoupling from third-party libraries, requirement specification, testing, deployment, etc.

\subsection{Academic and scientific writing}

One application area of ChatGPT that has garnered the interest of many is language summarization and elaboration. ChatGPT has been widely used for essay writing, application drafting, email content generation, and research paper writing since its inception. To that end, \cite{Biswas2023} has exploited ChatGPT to write an article related to medical content and has argued that the future of medical writing will be dependent on AI-assisted tools. Similarly, \cite{koo2023importance} has emphasized the proper use of the tool for disciplined medical writing and discussed the underlying concerns for this. Furthermore, \cite{kitamura2023chatgpt} has indicated that AI-assisted tools are very helpful and can be instrumental for future medical content writing. That said, human judgment is imperative to corroborate ChatGPT's output. In a similar vein,  \cite{kumar2023analysis} has asserted that ChatGPT has great potential in research writing if mentored by humans.

In addition to this, \cite{bishop2023computer} has shown with a series of conversations with ChatGPT that the AI bot can write in human style and may copy authors' styles of writing too. The authors \cite{salvagno2023can} have arguably inferred the need for consensus on how to regulate the use of AI-assisted tools in academic writing as the use of chatbots in scientific writing presents ethical issues related to the risk of plagiarism, inaccuracies, and unequal accessibility. An experiment by \cite{Gao20231} used 50 abstracts from scientific journals and asked ChatGPT to generate abstracts based on the titles. The generated abstracts were then reviewed by plagiarism detectors and blinded human reviewers. 68\% of the generated abstracts were correctly identified as such, and 14\% of the real abstracts were mistakenly identified as generated by ChatGPT. Interestingly, human reviewers found it difficult to differentiate between the abstracts written by the chatbot and those written by humans \cite{alkaissi2023artificial}.

\subsection{Research and Education}

For many application areas where chatGPT is being investigated, research and educations are the most prominent.   
\cite{rahman2023chatgpt} has experimentally shown that ChatGPT can be used to solve both technical problems, such as engineering and computer programming, and non-technical problems, such as language and literature. However, they warned to be aware of its limitations such as bias and discrimination, privacy and security, misuse of technology, accountability, transparency, and social impact. In a similar work, \cite{Tlili2023}, the study on ChatGPT was conducted in three stages. The first stage showed that social media discourse is generally positive and enthusiastic about using ChatGPT in education. The second stage analyzed ChatGPT's impact on educational transformation, response quality, usefulness, personality and emotion, and ethics. In the third stage, user experiences in ten educational scenarios revealed issues such as cheating, honesty, privacy, and manipulation. The study's findings highlight the need for the responsible and safe adoption of chatbots, specifically ChatGPT, in education. Furthermore, The article by \cite{hong2023impact} argues that ChatGPT presents significant opportunities for teachers and education institutes to enhance language teaching and assessments, leading to more personalized learning experiences. On top of this, authors have identified that the technology also offers a potential mechanism for researchers to explore new areas in research and education.

The authors in \cite{megahed2023generative} have concluded from their study that ChatGPT performs well in structured tasks such as translating code and explaining well-known concepts but struggles with nuanced tasks such as explaining less familiar terms and creating code from scratch. They suggested that while the use of AI tools may increase efficiency and productivity, current results may be misleading and incorrect. In contrast, \cite{halaweh2023chatgpt} has suggested that integration of ChatGPT could be beneficial and has given the outline to use it more efficiently.

To that end, the researchers exploring ChatGPT for research and education-related applications recommend using AI-assisted tools with censored and careful usage. Therefore, generative AI models must be properly validated and used in combination with other methods in software process improvement to ensure accurate results.

\subsection{Environmental Science}
There are a few studies so far on the potential use of ChatGPT in environmental sciences. \cite{Rillig20233464} outlined the potential benefits and risks of this LLM tool. They argued that ChatGPT can help in streamlining the workflow of environmental research where environmentalists can focus more on designing experiments and developing new ideas rather than the quality of their writing. It will also allow non-English speaking countries to have greater representation in the field of environmental science, accelerating the pace of research in relevant environmental issues. \cite{zhu2023chatgpt} also made similar observations but raised a few concerns also. Given the various decision-making processes involved in environmental research, it is essential to exercise caution while integrating AI-enabled tools such as ChatGPT into them. This is particularly important when addressing environmental issues that have a significant impact on the welfare of society.

\par \cite{biswas2023potential} mentioned the use of ChatGPT to address global warming. According to him, environment researchers can leverage the capabilities of ChatGPT to analyze and interpret vast amounts of climate change data and subsequently predict climate shift patterns based on the analysis. Furthermore, ChatGPT can be employed to present complex climate change information to a broader audience in an easily comprehensible format. It has the potential to offer policy-makers pertinent information and recommendations to mitigate climate variations. By inputting data, ChatGPT can also generate climate scenarios that can aid in making informed decisions. \cite{rathore2023future} proposed ChatGPT enabled sustainable and environment-friendly textile manufacturing process. She argued that ChatGpt could help in production process optimization. Additionally, it could be employed to provide automated customer cell that is both relevant and valuable. Another possibility is the development of fine-tuned recommendations according to the needs and preferences of buyers.

\subsection{Natural Language Processing}
ChatGPT has shown its potential as a valuable tool for various NLP-oriented tasks, including suicide tendency detection, hate speech detection, and fake news detection \cite{amin2023will, qin2023chatgpt, hendy2023good}. In particular, \cite{qin2023chatgpt} argued that larger models like ChatGPT can perform NLP tasks without the need for specific data adaptation. They conducted an evaluation of ChatGPT's zero-shot learning ability on 20 common NLP datasets, covering categories such as reasoning, natural language inference, question answering, dialogue, summarization, named entity recognition, and sentiment analysis. The evaluation results indicated that ChatGPT excelled in tasks that required reasoning skills, including arithmetic reasoning while facing challenges in tasks such as sequence tagging.

In their study, Hendy et al. \cite{hendy2023good} conducted a comprehensive evaluation of GPT models, including ChatGPT, for machine translation tasks. The evaluation covered 18 translation directions involving a diverse range of languages such as French, German, Icelandic, Chinese, Japanese, and others. The results demonstrated that GPT models could generate translation outputs that were highly competitive for languages with abundant resources. However, for low-resource languages, the current state of GPT models exhibited limitations, indicating the need for further improvements.
Similarly, Amin et al. \cite{amin2023will} conducted an analysis of various NLP tasks, including suicide tendency detection and personality prediction. They compared the performance of ChatGPT with both simple and sophisticated models like RoBERTa. The findings revealed that task-specific models like RoBERTa outperformed ChatGPT, particularly in specialized downstream tasks. However, ChatGPT still demonstrated satisfactory performance compared to baseline models in a variety of tasks.

Given the similarities among various NLP tasks, the findings from the aforementioned studies can be extrapolated to other related areas, including the assessment of news article accuracy, especially in the detection of fake news. Additionally, the application of ChatGPT is rapidly expanding across various domains \cite{Mijwil202365,aljanabi2023chatgpt}, indicating a growing trend that is expected to continue in the foreseeable future.

\section{Challenges and issues of ChatGPT}
Researchers have identified several issues regarding ChatGPT, which can be broadly categorized into two groups: intrinsic limitations and usage-related concerns. These categories, along with their respective limitations, are presented in Table \ref{tab:limitations} for better clarity and understanding. These limitations make the usage and deployment of ChatGPT difficult in real-world scenarios.

\subsection{Intrinsic}
 Intrinsic issues refer to the limitations inherent to ChatGPT and can be overcome primarily by the tool's developers through algorithm enhancements and/or upgraded training data. It includes five major limitations, namely, hallucination, biased content, not real-time, misinformation and inexplicability.  ChatGPT might hallucinate i.e., create new data/information which does not exist \cite{deng2022benefits}. Another similar concern is misinformation \cite{borji2023categorical}. Both issues can lead to the creation of counterfactual or meaningless responses, which can seriously threaten the reliability of the generated content. The false narratives generated by ChatGPT can be easily mistaken as legitimate, particularly by individuals unfamiliar with the subject matter \cite{chatGPTFake}. Algorithmic improvement, inputting the queries properly, and verifying generated responses might help overcome these problems. Reinforcement learning through human feedback will also help ChatGPT to improve the factuality of its responses \cite{stiennon2020learning}. 

\par There are also concerns regarding potential harms related to stereotypes and biased responses caused by ChatGPT \cite{liang2021towards,nadeem2020stereoset}. Besides algorithmic improvement and human feedback, refining the training data to remove or mark the biased content might help in this direction. There are many critical applications of ChatGPT where sound reasoning and explanation of logical deduction steps are required. It includes decision-making in various fault-intolerant domains such as financial services, environmental sciences, healthcare, etc. In such scenarios, ChatGPT must not only provide accurate information that can be used for decision-making, the steps involved in the logical reasoning deduction process also be mentioned \cite{ wei2022chain}.

\begin{table*}
\scriptsize
\centering
\caption{Major issues of ChatGPT and their potential solutions}
\begin{tabular}{|l|l|l|l|l|}
\hline
\textbf{Category}                                                         & \textbf{Issue}      & \textbf{Description}                                                                                                               & \textbf{Potential solution(s)}                                                                                                                                                                          & \textbf{References}                                                                                                                 \\ \hline
\multirow{5}{*}{Inherent}                                                 & Hallucination       & \begin{tabular}[c]{@{}l@{}}creating new data/information \\ which does not exist\end{tabular}                                      & \begin{tabular}[c]{@{}l@{}}Algorithmic   improvement; Users should \\ use appropriate prompts; Verifying the \\ generated content\end{tabular}                                                          & \begin{tabular}[c]{@{}l@{}}\cite{deng2022benefits}\\  \cite{cao2023comprehensive}\end{tabular}  \\ \cline{2-5} 
                                                                          & Biased   content    & \begin{tabular}[c]{@{}l@{}}producing negative comments/\\ generalizations related to race, \\ religion, gender, etc.\end{tabular}  & \begin{tabular}[c]{@{}l@{}}Algorithmic improvement; Refining the \\ training data, feedback by humans in case \\ of biased content\end{tabular}                                                         & \begin{tabular}[c]{@{}l@{}}\cite{liang2021towards}\\ \cite{nadeem2020stereoset}\end{tabular} \\ \cline{2-5} 
                                                                          & Not   real-time     & \begin{tabular}[c]{@{}l@{}}does   not has access to \\ real-time information as it \\ was fed information till 2021.\end{tabular}  & \begin{tabular}[c]{@{}l@{}}Providing direct access to real-time/online \\ data; Algorithm improvement and speed-up \\ are required\end{tabular}                                                         & \begin{tabular}[c]{@{}l@{}}\cite{chatGPT}\\ \cite{gpt4}\end{tabular}                            \\ \cline{2-5} 
                                                                          & Misinformation      & \begin{tabular}[c]{@{}l@{}}generating factually incorrect \\ information.\end{tabular}                                             & \begin{tabular}[c]{@{}l@{}}Designing factuality-based measures to \\ indicate the level of misinformation; \\ Mentioning the references; Tagging by \\ humans in case of false information\end{tabular} & \cite{borji2023categorical}                                                                                       \\ \cline{2-5} 
                                                                          & Inexplicability     & \begin{tabular}[c]{@{}l@{}}not explaining the steps of \\ information generation in \\ critical decision-making tasks\end{tabular} & \begin{tabular}[c]{@{}l@{}}Mentioning the steps involved in \\ logical reasoning deduction process\end{tabular}                                                                                         & \begin{tabular}[c]{@{}l@{}}\cite{cao2023comprehensive}\\ \cite{chatGPTFake}\end{tabular}        \\ \hline
\multirow{3}{*}{\begin{tabular}[c]{@{}l@{}}Usage \\ related\end{tabular}} & Ethical   issues    & \begin{tabular}[c]{@{}l@{}}not acknowledging ChatGPT \\ whenever content is generated\\ using it.\end{tabular}                     & \begin{tabular}[c]{@{}l@{}}mentioning ChatGPT as author/source \\ of information; Laws should be designed\\  to avoid unethical usage of ChatGPT\end{tabular}                                           & \begin{tabular}[c]{@{}l@{}}\cite{Elali2023}\\  \cite{cao2023comprehensive}\end{tabular}          \\ \cline{2-5} 
                                                                          & Copyright violation & \begin{tabular}[c]{@{}l@{}}generating full/partial content \\ identical to previous works \\ without prior consent\end{tabular}    & \begin{tabular}[c]{@{}l@{}}Verifying the generated content before\\  using/publishing;  Tagging by humans\\  in case of copyright violation\end{tabular}                                                & \cite{sallam2023utility}                                                                                          \\ \cline{2-5} 
                                                                          & Over-reliance       & \begin{tabular}[c]{@{}l@{}}may make humans lazy and \\ apathetic and always rely on\\  the generated information\end{tabular}      & \begin{tabular}[c]{@{}l@{}}Humans should verify the generated content \\ and use ChatGPT as a tool only to produce \\ better outcomes.\end{tabular}                                                     & \cite{sallam2023chatgpt}                                                                                          \\ \hline
\end{tabular}
\label{tab:limitations}
\end{table*}

 \subsection{Usage-related}
 The category of usage-related issues includes unethical usage of the tool, copyright-infringed content, and over-reliance on ChatGPT. Ethical concerns arise, particularly when the tool is used to generate the content without acknowledgment. Unethical use also includes the deliberate generation of manipulated content that can promote misinformation and provoke violence, creating damage at an individual or organizational level \cite{Elali2023,cao2023comprehensive}. An ethical usage of the tool includes mentioning ChatGPT as the author/source of the generated information. In fact, few publishers have recognized the use of ChatGPT for academic writing to promote its ethical usage practices \cite{ chatGPTUse}. Also, laws and regulations should be designed to penalize the unethical usage of ChatGPT. Since there are many ethical considerations related to ChatGPT, we have discussed them separately in the next section. There are also concerns that ChatGPT-generated content may lead to violations of copyright and intellectual property rights \cite{ sallam2023utility}. Copyright infringement primarily includes generating full/partial information identical to already published works without the prior consent of the owner. Since ChatGPT is unaware of copyright materials, verifying the generated content before using/publishing and tagging by humans in case of copyright violation might help to resolve this issue. Lastly, there is a fear of over-reliance on the tool, which may make humans lazy and apathetic and make them always rely on the generated information \cite{sallam2023chatgpt}. Therefore, we should verify the generated content and use ChatGPT as a tool only to produce better outcomes.

\subsection{Ethical concerns}
ChatGPT has the capability to automatically generate responses by drawing information from numerous internet sources, often without requiring further input from the user. This has raised concerns regarding its potential misuse, as individuals have reportedly utilized the system to create university essays and scholarly articles, even including references if prompted \cite{ali2023readership}. One of the ethical issues associated with the usage of ChatGPT is the generation of fake text and narratives \cite{chatGPTFake,dugan2022real}. It is worth noting that the detection of artificially generated fake text and meaningless information is not a new challenge \cite{amancio2015comparing,jawahar2020automatic}. The identification of fake text generated by ChatGPT can be viewed as a two-step process. The first step involves determining whether a given text is created through ChatGPT. Once this is established, the second step requires identifying whether the text itself is fake or genuine. The latter process, which is fake text identification, falls within a well-established domain, with many state-of-the-art algorithms and techniques available to tag fake text with significant accuracy \cite{zhou2020survey}.
However, the former step, which focuses on specifically identifying texts generated by ChatGPT, is relatively new, and researchers are actively working to address this issue \cite{dugan2022real, mitchell2023detectgpt, Curtis2023275,mitrovic2023chatgpt}.

Curtis et al. \cite{Curtis2023275} have proposed various methods for identifying text generated by ChatGPT. These approaches encompass simple binary classifiers as well as advanced deep-learning models. Some techniques leverage statistical characteristics or syntactic patterns, while others incorporate semantic and contextual information to enhance accuracy. The primary objective of these studies was to provide a comprehensive and current evaluation of the most recent detection techniques specific to ChatGPT. Additionally, they assessed the performance of other AI-generated text detection tools that were not specifically designed for ChatGPT-generated content.
In a different study, \cite{mitrovic2023chatgpt} focused on brief online reviews and conducted two experiments to compare text generated by humans and ChatGPT. In the first experiment, they generated ChatGPT text using custom queries, while in the second experiment, they rephrased original human-generated reviews to obtain alternative text. They fine-tuned a model based on the Transformer architecture and employed it for making predictions. By comparing their model with an approach based on perplexity scores, they found that differentiating between human-generated and ChatGPT-generated reviews is more challenging for the machine-learning model when using rephrased text.

The authors in \cite{gao2023comparing} collected a total of 50 research abstracts by selecting ten abstracts from five prestigious medical journals with high impact factors. They used ChatGPT to generate research abstracts by providing titles and journal names as prompts. To evaluate the quality and authenticity of the abstracts, they employed an artificial intelligence (AI) output detector, a plagiarism detector and involved human reviewers who were unaware of the origin of the abstracts. These reviewers determined whether the abstracts were original or generated. Moving on, Mitchell et al. \cite{mitchell2023detectgpt} proposed a novel criterion called DetectGPT, which relied on curvature to determine if a passage was generated from a specific Language Model (LM) such as GPT. Unlike other methods, DetectGPT did not require training a separate classifier, creating a dataset of real or generated passages, or applying explicit watermarks to the generated text. Instead, it relied solely on log probabilities calculated by the target model and introduced random perturbations to the passage using a different generic pre-trained language model.

The increasing use of ChatGPT underscores the pressing need for rigorous AI author guidelines in academic publishing. There are ethical concerns related to copyright, attribution, plagiarism, and authorship when AI generates academic text. These concerns are particularly relevant because current technology does not allow human readers or anti-plagiarism software to distinguish between AI-generated and human-authored content \cite{rahimi2023chatgpt}. While some studies have credited ChatGPT as an author, there is an ongoing debate about whether generative AI meets the International Committee of Medical Journal Editors' authorship criteria. Can a chatbot truly provide approval for work and be held accountable for its content? The Committee on Publication Ethics and the International Association of Scientific, Technical, and Medical Publishers have developed AI recommendations for editorial decision-making and ethics, respectively \cite{zhuo2023exploring}. As AI technology becomes more tailored to user needs and more widely used, we believe it is crucial to have comprehensive discussions about authorship policies. Major publishers like Elsevier, who publish the Lancet family of journals, have already stated that AI cannot be listed as an author and that its use must be properly acknowledged \cite{sallam2023chatgpt}.

Our view is that ChatGPT's availability, ease of use, and multi-language capabilities could significantly boost scholarly output, thereby democratizing knowledge dissemination. However, the chatbot's potential to generate misleading or inaccurate content raises concerns about scholarly misinformation \cite{mhlanga2023open}. As demonstrated by the COVID-19 infodemic, the spread of misinformation in medical publishing can have serious societal consequences. OpenAI has acknowledged that ChatGPT may produce plausible-sounding yet incorrect or nonsensical answers \cite{lund2023chatgpt}.

\section{Future possibilities}
In this section, we explore some of the future possibilities related to ChatGPT. We envision that future iterations of ChatGPT might incorporate various additional variables, which can help develop a more sophisticated and enhanced AI language model.

\subsection{Improving Conversational Capabilities}
\begin{figure*}[t!]
\centering
\includegraphics[width=1.0\linewidth]{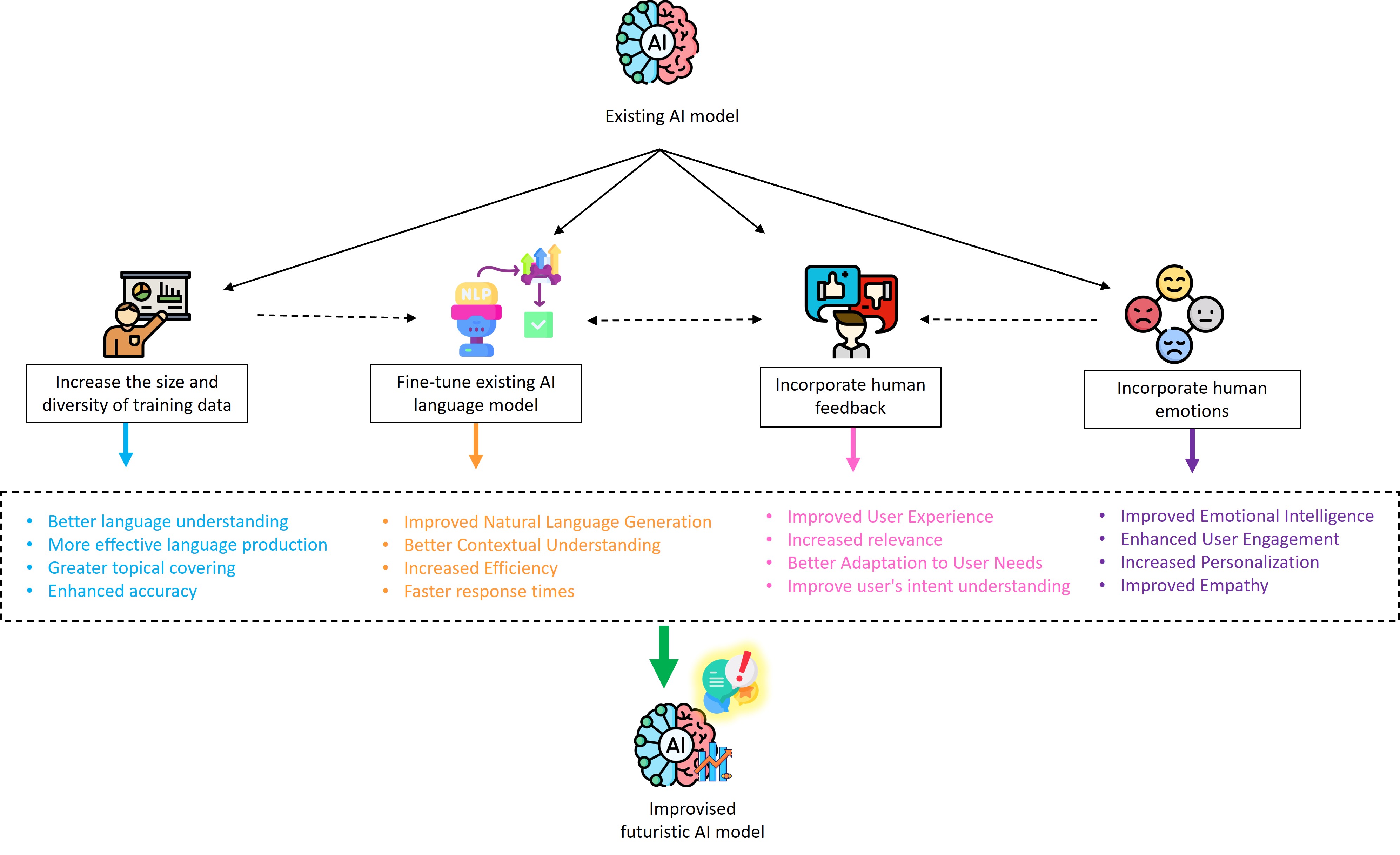}
\caption{Enhancing the conversational ability of ChatGPT}
\label{fig:ImprovedChatGPT}
\end{figure*}

ChatGPT may become even better at comprehending and reacting to human speech, making it sound more conversational and natural. This might entail developments in disciplines like sentiment analysis, natural language processing, and contextual comprehension. The following are some basic strategies that could help AI become more conversational (Figure \ref{fig:ImprovedChatGPT}).

\subsubsection{Increasing the volume and variety of training data}
AI language models learn from the provided data. By exposing them to a wider variety of linguistic patterns and linguistic contexts, expanding the size and diversity of the training data can therefore aid in the improvement of their conversational abilities.
Typically, increasing the volume and variety of training data can help improve ChatGPT's performance. ChatGPT is a machine learning model that uses a large amount of training data to learn patterns and make predictions based on them. By providing more data, the model can improve its understanding of language and the relationships between words and phrases \cite{cao2023comprehensive}.
More specifically, increasing the volume of training data can help the model learn more about different topics and contexts, which can make it more versatile and able to handle a wider range of queries. Additionally, providing more varied data can help the model learn to recognize and understand different types of language and speech patterns, improving its ability to handle diverse inputs and generate accurate responses.

However, it is also important to note that simply increasing the volume and variety of data alone may not always result in improved performance. The quality of the data is also crucial, and it is important to ensure that the data used for training is accurate, relevant, and diverse enough to represent the full range of language and speech patterns. Additionally, other factors such as the model architecture, training methods, and hyperparameters can also impact the model's performance.

\subsubsection{Fine-tuning}
The method of fine-tuning involves putting an existing AI language model through a series of tasks or domains. The model can be trained to produce more pertinent and useful answers by honing in on conversational tasks like customer service or personal assistants \cite{bakker2022fine,himeur2022next}.
In the case of ChatGPT, fine-tuning allows the model to learn and understand the nuances of natural language conversations, enabling it to generate more human-like responses. This process involves feeding the model with conversational datasets and optimizing it through backpropagation using the conversation pairs as input and output. As a result, the model becomes more accurate, efficient, and responsive to the specific task of generating conversational responses \cite{dwivedi2023so,hoppner2023chatgpt}.

\subsubsection{Incorporating human feedback}
Human input on the responses produced by AI language models can be gathered to assist the models' conversational skills. This can be achieved by either asking users to rate the quality of the answers or by having humans review, edit, and offer feedback on the responses produced by the model. Over time, it would help people grasp the conversation's context better and respond appropriately. The models can produce more pertinent and personalized responses if their ability to comprehend contexts, such as prior conversation history or the user's intent, is improved.
Incorporating human feedback is an effective way to improve the performance of ChatGPT. The first step is to collect feedback from users about the generated responses. One way to do this is to provide users with the option to rate the quality of the responses or provide suggestions for improvement.
Once the feedback is collected, it needs to be analyzed to identify the areas of improvement. This can be done using NLP techniques to extract relevant information and classify the feedback into different categories. The next step is to incorporate the feedback into the ChatGPT model. One way to do this is to use reinforcement learning, which involves modifying the model's weights to maximize a reward signal based on user feedback. Another way is to use the feedback to fine-tune the model and retrain it on the specific areas that require improvement. After incorporating the feedback, it's essential to evaluate the performance of the updated ChatGPT model. This can be done by measuring the quality of the responses generated by the model and comparing it to the previous version. Finally, the process of collecting feedback, analyzing it, and incorporating it into the model needs to be repeated iteratively to ensure continued improvement in the ChatGPT model's performance.

\subsubsection{Incorporating human emotions}
Humans frequently express their feelings through humor, empathy, and sarcasm. AI language models' conversational skills can be enhanced by adding feelings to make them more relatable and interesting. Although it is a complex and debated subject in the area of artificial intelligence \cite{Pahl2022FemaleW2}; \cite{Domnich2021ResponsibleAG}, incorporating human emotions into language models for AI is a challenge. Although feelings are a vital aspect of human communication, they can also be unpredictable and influenced by personal perspectives. As a result, adding feelings to AI language models runs the risk of unintended consequences and risks like bias and discrimination, manipulation, privacy evasion, and inappropriate or offensive responses. Therefore, before including emotions in AI language models, it is crucial to thoroughly consider the risks and ethical ramifications. When creating and implementing these kinds of systems, developers should prioritize accountability, transparency, and user permission. To make sure that the possible risks and benefits are completely comprehended and mitigated, it is also crucial to involve a variety of stakeholders, including experts on psychology and ethics.

\subsubsection{Style-based technique for higher level text analysis}

By combining style-based techniques that utilize complex networks with Chat GPT, the model can leverage the analysis of stylistic attributes to enhance its text generation capabilities. This combination allows Chat GPT to identify, understand, and replicate stylistic patterns, resulting in text that aligns not only with the content but also with the desired style. The integration of style-based analysis within the Chat GPT framework enables a more comprehensive understanding and generation of text, opening up possibilities for personalized, stylistically rich, and contextually appropriate interactions. For instance, \cite{correa2019word} has introduced a method to induce word senses by leveraging word embeddings and community detection in complex networks. Chat GPT can potentially provide an innovative extension by incorporating style metrics and adequate prompting techniques.

In addition to this, Chat GPT can contribute innovatively by integrating word embedding \cite{quispe2021using} to enhance its understanding and incorporation of style elements. By leveraging virtual edges and considering stylistic features, Chat GPT can generate text that exhibits desired stylistic nuances. Furthermore, integrating style-based text analysis into Chat GPT's conversational framework \cite{stella2019forma} can help in generating text that feels more natural and personalized, fostering a stronger connection between the user and the AI system. However, the main challenge at hand is to adeptly merge Chat GPT's capacity for generating content-focused text with the style analysis facilitated by complex network-based approaches. It requires seamless integration of Chat GPT's content generation prowess with the comprehensive style analysis offered by complex network techniques. Achieving this synergy is vital to effectively combine the strengths of both approaches and produce text that not only captures the intended content but also reflects the desired stylistic attributes. Arguably, effective conversational AI systems like ChatGPT require a combination of various techniques and considerations to deliver a satisfying and human-like conversational experience, for example, considering factors such as context understanding, coherence, and response relevance.

\subsection{Personalization}
\begin{figure*}[t!]
\centering
\includegraphics[width=1.0\linewidth]{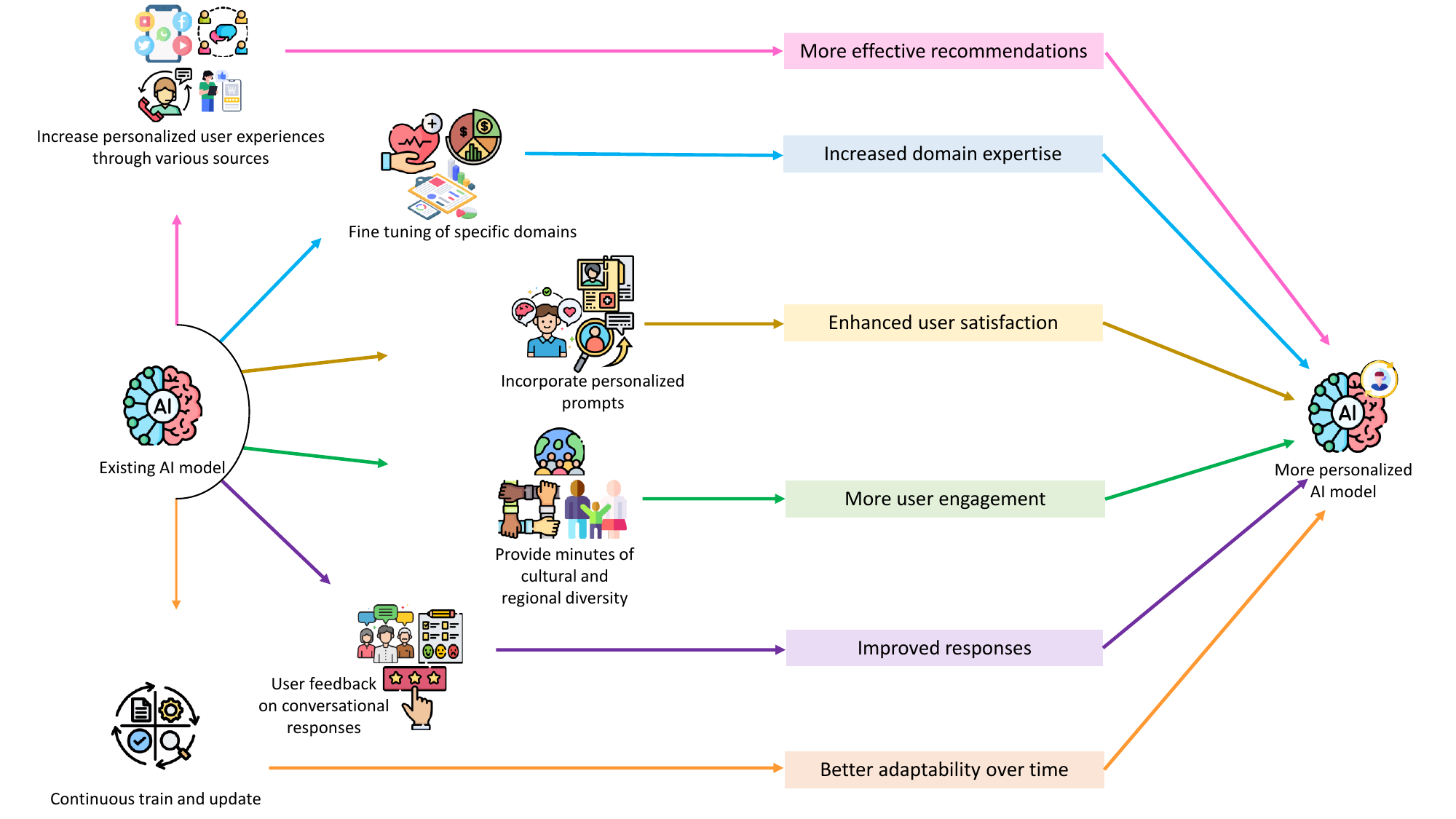}
\caption{\textcolor{black}{Domain/person-specific personalization of ChatGPT.}}
\label{fig:personalizedChatGPT}
\end{figure*}

Future iterations of ChatGPT might be adapted to each user, making use of their prior interactions to personalize answers and create more intimate conversations. User profiling and data protection could both be improved as a result. Important considerations for enhancing ChatGPT's customization abilities are as follows (Figure \ref{fig:personalizedChatGPT}).

\subsubsection{Increase personalized user experiences through various sources}
More information can be provided to improve understanding of linguistic patterns and enable answers to be more user-specific. Numerous sources, including social media, customer support interactions, and other online conversations, can provide this information. More diverse data and a deeper comprehension of linguistic nuance would ultimately lead to more personalized recommendations and responses \cite{dwivedi2021setting}.

\subsubsection{Fine-tuning of specific domains}
Increasing domain knowledge in a particular topic or domain, such as customer service, healthcare, business, or finance, can be accomplished by fine-tuning a particular dataset. For users in that particular domain, this may result in answers that are more precise and tailored \cite{batko2022use,himeur2022ai}. For instance, if we wanted to increase ChatGPT's capacity to offer tailored responses to customers in the medical field, we could do so by including more information about medical history and contemporary medical advancements in blogs, social media platforms, and periodicals. With the aid of this information, ChatGPT will be better able to provide customers in the medical industry with personalized answers by better comprehending the linguistic conventions and terminologies used there \cite{eysenbach2023role,kung2023performance}.

\subsubsection{Incorporating personalized prompts}
Including personalized prompts, such as the user's name or reference to prior conversations, can enhance user satisfaction and improve understanding. ChatGPT can use a user's name in the answer if it is known, making the interaction more customized. For instance, ChatGPT might reply. For instance, let's consider a scenario where a user named John has engaged in a previous conversation with ChatGPT, during which he mentioned his name. If John were to inquire, "What's the weather like today?" ChatGPT could respond with, "Certainly, John! Today's forecast is sunny with a temperature of 75 degrees." By incorporating John's name and previous conversation, ChatGPT is capable of delivering a more personalized and customized response, thereby enhancing the user's overall satisfaction and comprehension of the information provided \cite{jungwirth2023artificial}.

\subsubsection{Provide instances of cultural and regional diversity}
ChatGPT can be trained on a diverse collection of data that includes details about various cultural norms and traditions, such as greetings, social customs, and cultural practices. This can aid ChatGPT in better comprehending and addressing users from various cultural backgrounds and ensuring that its retorts are considerate and respectful of various cultural norms and standards.

\subsubsection{User feedback on conversational responses}
Gathering user feedback on conversational responses can help pinpoint ways to make them more individualized. Surveys, user trials, and analysis of user interactions with the model are all ways to gather feedback. This can direct upcoming updates and training while also pointing out areas where the model needs development. Users' comments on ChatGPT can be gathered to determine what needs to be improved. As an illustration, if a user asks a query and ChatGPT responds incorrectly, the user can offer feedback to fix the error. With time, ChatGPT can use this input to enhance the precision and customization of its responses.

\subsubsection{Continuous training and updating}
In order to enhance ChatGPT's capacity to deliver personalized answers, new data can be trained on it. For instance, if a new fashion trend emerges, ChatGPT can be trained on data pertaining to that trend to enhance its capacity to offer individualized answers about it. 
Specifically, there are several ways to achieve continuous training and updating of ChatGPT. One approach is to feed new data into the model on a regular basis, either by adding new data to the existing training set or by fine-tuning the model on new data. This can be done through techniques such as transfer learning, where the model is trained on a large dataset and then fine-tuned on a smaller, more specific dataset.
Another approach is to continually monitor the performance of ChatGPT and make adjustments to the model as needed. This can involve monitoring metrics such as accuracy, perplexity, and language generation quality and using this feedback to update the model's architecture or training process.
Additionally, it is important to note that continuous training and updating of ChatGPT requires ongoing resources and infrastructure to support it. This includes access to large amounts of training data, computing power, and storage space to store the model and its updates.

\subsection{Multimodal design}
\begin{figure*}[t!]
\centering
\includegraphics[width=1.0\linewidth]{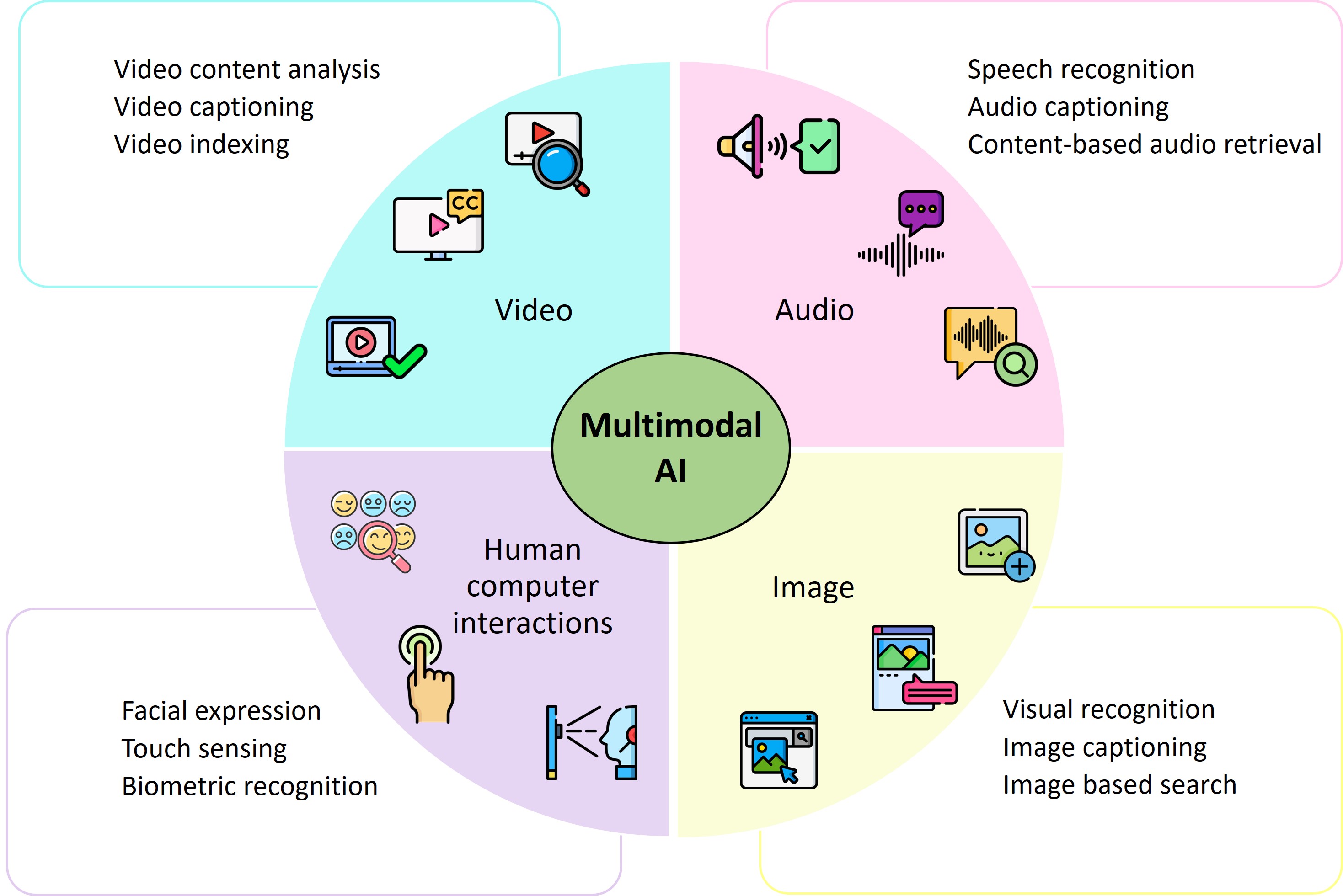}
\caption{Various technological integration for building multimodal AI}
\label{fig:Multimodal AI}
\end{figure*}

Multimodal integration to ChatGPT would enable more natural and human-like intuitive, engaging, and effective communication. It entails creating machine learning models and algorithms capable of processing and combining a variety of data, including text, audio, and images. These different modalities can be combined and integrated into various ways to create more engaging and effective user experiences. Some key aspects of designing multimodal AI are as follows (Figure \ref{fig:Multimodal AI}).

\subsubsection{Image-based design} 
The use of images as the main tool for communicating is the focus of image-based design. In order to convey a certain message or idea, this can involve the use of images, illustrations, and other visual components. Visual recognition, image captioning, and image-based search can be integrated with ChatGPT to make it a multimodal AI. For instance, image recognition and analysis can be used to determine the most useful visual aspects to employ in a design, and image-based search can be used to find relevant visual content. Similarly, image captioning can be used to add descriptive or explanatory text to images, which can help to make them more accessible and informative for viewers. Image recognition and image-based search are powerful tools that have numerous applications in different fields. For instance, students can utilize these technologies to locate and analyze relevant images for academic research purposes. On the other hand, medical professionals can use image recognition to examine medical images such as X-rays and ECG graphs, which can help with the diagnosis of various conditions. Additionally, image-based search can be utilized to find relevant medical images for research purposes. Moreover, Image captioning can be used to add descriptive text to product photos, making them more engaging and informative for potential customers. This can be especially helpful for users who are visually impaired and rely on screen readers to access content.

\subsubsection{Audio-based design} 
A few examples of audio features that can be incorporated with ChatGPT are speech recognition, audio captioning, and content-based audio retrieval. By processing aural input and translating it into text that software or other devices may use, speech recognition technology enables ChatGPT to comprehend and recognize spoken words. It would be beneficial for persons who are deaf or hard of hearing, as well as non-native speakers, to add text subtitles to audio information or audio captioning because it can help make audio content more accessible. Audio subtitles for audio content can be produced manually or automatically using voice recognition software. In addition, users could search for audio content based on the characteristics of that content using content-based audio retrieval.

\subsubsection{Video-based design} 
Integrating video-based design technologies such as video content analysis, video captioning, and video indexing is a crucial aspect of developing a multimodal AI system. By incorporating video content analysis, a multimodal ChatGPT can analyze and process video footage, detect objects, and track motions within the video. Video captioning can be added to ChatGPT to provide closed captions or subtitles during conversations, making the video content more accessible. Additionally, video indexing allows users to search for specific content within a video using keywords or timestamps, making it easier to find relevant information quickly. This makes video-based design technologies essential for developing a robust and effective multimodal AI system.

\subsubsection{Human-computer interaction design}
The next step towards creating a multimodal AI would be to incorporate characteristics of human-computer interaction including facial expressions, touch sensitivity, and biometric recognition. For instance, face recognition technology would enable ChatGPT to recognize and react to various facial emotions. A more natural and tactile experience for the user can be achieved by adding touch sensitivity to ChatGPT, allowing it to recognize and react to touch inputs. By incorporating biometric authentication technology, such as fingerprint or iris scanning, ChatGPT may offer improved security and user authentication, making it a more dependable and trustworthy platform for critical activities.

\subsection{Trustworthiness}
\begin{figure*}[t!]
\centering
\includegraphics[width=1.0\linewidth]{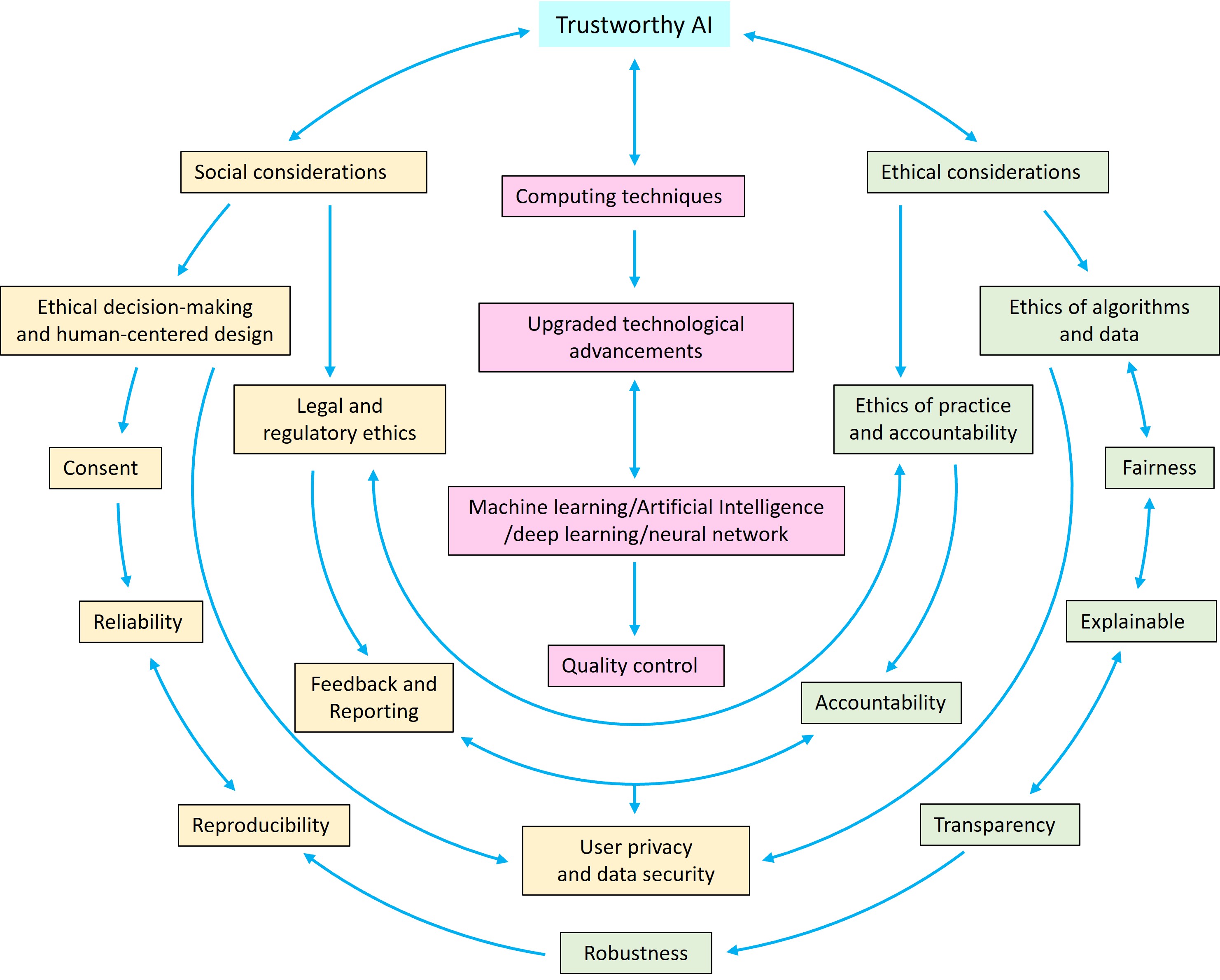}
\caption{Interplay of various factors to achieve Trustworthy AI.}
\label{fig:TrustworthyAI}
\end{figure*}

One of the most pressing needs today is the development of trustworthy AI. To achieve this, future iterations of ChatGPT could incorporate features that guarantee impartial and fair answers. As AI ethics and fairness become more critical, three key categories could be considered: computing techniques, ethical considerations, and social considerations.
Upgraded technological developments such as deep learning, machine learning, and artificial neural networks should be integrated into computing methods to improve AI performance. However, ethical considerations must also be taken into account, including data ethics, to ensure that data gathering, storage, and use for AI system training are conducted ethically and responsibly while being protected from unauthorized access or abuse.
To prevent discrimination against individuals or groups based on characteristics such as religion, ethnicity, or gender, machine learning fairness must also be a priority. Privacy protection is another crucial factor that involves strategies such as encryption and differential privacy to safeguard user data and ensure their privacy is not violated. These measures are crucial for fostering user confidence and ensuring that AI operates in an ethical and moral manner \cite{rahimi2023chatgpt}.
To illustrate the factors related to ensuring trust in AI tools, a detailed diagram is presented in Figure \ref{fig:TrustworthyAI}.

\par As an artificial intelligence language model (\cite{Abdel-Messih2023}), ChatGPT is made to produce responses based on the recurring patterns and relationships in the incoming data that it has been trained on. Future iterations of ChatGPT should include features to guarantee impartial and fair responses, though, in order to create a trustworthy AI. The inclusion of a wider variety of sources and viewpoints as well as efforts to correct any biases that may be present in the data can all help to enhance training data. One way to improve training data is to resolve biases and include a wider variety of sources and viewpoints. This can lessen the chance that ChatGPT's answers will reinforce preexisting biases or stereotypes. Fairness and responsibility can also be supported by transparency in the decision-making process. A transparent and explicable ChatGPT would offer detailed justifications for the choices it makes and the reasons behind the responses it generates. This might entail informing users of the characteristics or elements that go into producing a response. Incorporating methods for monitoring and feedback can also aid in identifying and resolving any potential problems. Encouraging fairness and moral conduct can help guarantee that ChatGPT keeps developing and getting better over time. To guarantee that ChatGPT is producing responses that are in line with moral and ethical standards, regular auditing and monitoring of the decision-making process is required. By implementing these suggestions, ChatGPT can develop into an AI that users can trust to produce true, objective, and moral answers. Some of the key components in building a trustworthy AI:
 
\subsubsection{Fairness}
Fairness in AI is should be incorporated to avoid biases and discrimination during the creation and application of AI systems. It entails ensuring that AI systems handle everyone equally and without discrimination and do not reinforce existing biases and inequalities. This requires a thorough consideration of the data used to train AI models, as well as the algorithms and decision-making procedures used in AI systems.
As an AI language model, ChatGPT can strive to achieve fairness by considering and addressing bias in its training data, decision-making processes, and outputs. One way to address bias in training data is to ensure that the data used to train the model is diverse and representative of the population. This can be achieved by using a variety of sources and by including data from underrepresented groups \cite{hassani2023role}. The training data should also be carefully curated and filtered to remove any biased or problematic data. Additionally, ChatGPT can incorporate fairness into its decision-making processes by using algorithms that account for fairness, such as those that use counterfactual reasoning or causal inference. These algorithms can help identify and correct potential biases in the model's outputs.
Finally, ChatGPT can strive to achieve fairness in its outputs by monitoring and analyzing its results to ensure that they are not unfairly biased against certain groups. If biases are detected, the model can be adjusted and retrained to address these issues.

Overall, It is important to note that achieving fairness is an ongoing process that requires continuous monitoring and improvement. As such, ChatGPT can also work to ensure transparency and accountability by providing explanations for its decisions and allowing for user feedback and oversight.

\subsubsection{Transparency}
The degree to which an AI system's decision-making processes and fundamental data are transparent and available to users is referred to as transparency in AI. Because transparent AI systems are simpler to comprehend and analyze, they encourage trust and make it more likely that the system will make ethical and legal choices.
As an AI language model, ChatGPT is designed to generate responses based on the input it receives. While it can provide information and insights, it does not have the ability to actively disclose information about itself or the data it uses to generate responses. However, to ensure transparency in the use of ChatGPT, it is important to be clear about the limitations of the model and its capabilities. Additionally, it is important to be transparent about the data sources used to train the model and any potential biases. To further enhance transparency, it may be useful to provide users with information about how the model is being used, such as in what contexts it is being deployed and how it is being monitored to ensure accuracy and fairness.

\subsubsection{Explainability}
An explainable AI could provide a comprehensive justification for its choices and decisions. As a result, humans can comprehend and validate the system's decision-making process, which is crucial for building confidence in AI systems.
ChatGPT is a machine learning model whose responses are based on statistical patterns in the data it was trained on. As such, it may not always provide the most nuanced or complete answer to a question, and it may occasionally make mistakes or provide incorrect information.
One approach to increasing explainability is to use models that are inherently interpretable, such as decision trees or linear regression models. These models are easier to understand because they explicitly show how each feature contributes to the model's output.

Another method is to calculate and present feature importance scores for the model's inputs. This allows users to see which factors the model is using to make its predictions.
Moving forward, creating visualizations can help users understand how the model is working. This might include showing which inputs are most influential, how the model is making decisions, and how it is arriving at its final output.
Moreover, adding a human-in-the-loop approach can also help increase the explainability of AI models. For example, providing users with the ability to ask follow-up questions about the model's predictions can help them better understand how the model arrived at its decision. Lastly,  providing clear and concise documentation that describes how the model was trained, what data was used, and how it is intended to be used can also increase the transparency and explainability of the AI model.


\subsubsection{Human-centered design}
It refers to the process of creating AI systems that are in tune with human values, needs, and preferences. This entails considering the ethical and social effects of AI systems and designing them with an emphasis on transparency, fairness, and accountability. This may necessitate incorporating moral principles into the design process, such as the principles of beneficence (doing good), non-maleficence (avoiding harm), and respect for values. 
In the case of developing AI language models, including ChatGPT, Human-Centered Design principles are usually applied in the research and development process, such as conducting user research to understand the needs and behaviors of people who use language models, testing different design prototypes, and iterating on the design based on user feedback.
\begin{figure*}
\begin{center}
\begin{tabular}{|m{16cm}|}
\hline
{\small \textbf{Abbreviations:}} \\ 
\begin{multicols}{2}
\footnotesize
\begin{acronym}[AWGN] 
\acro{AI}{artificial intelligence}
\acro{AIGC}{Artificial Intelligence Generated Content}
\acro{BERT}{Bidirectional Encoder Representations from Transformers}
\acro{ChatGPT}{Chat Generative Pre-trained Transformer} 
\acro{CoCoNuT}{Combining Context-aware Neural Translation models}
\acro{Codex}{GPT language model finetuned on publicly available code}
\acro{COVID}{Coronavirus Disease}
\acro{ECG}{Electroardiogram: a test to check heart's rhythm}
\acro{GAI}{Generative Artificial Intelligence}
\acro{GPT}{Generative Pre-trained Transformer}
\acro{LLMs}{Large language models}
\acro{NLP}{Natural Language Processing}
\acro{RLHF}{Reinforcement Learning from Human Feedback}
\acro{RoBERTa}{Robustly Optimized BERT Pretraining}
\end{acronym}
\end{multicols} \\
\hline
\end{tabular}
\end{center}
\end{figure*}

\section{Conclusion}
In this review article, we have showcased the immense potential of the future GPT language model in various fields by comprehensively reviewing more than 100 Scopus-indexed publications on ChatGPT. A novel taxonomy has been put forth which reveals a diverse range of applications across various domains such as healthcare, marketing \& finance, environment, education \& research, and academic writing. Despite its potential, the early ChatGPT researches still face some limitations. We have identified certain issues that may need to be addressed, which are classified as intrinsic and usage-centric. Additionally, we have identified and discussed ethical concerns. To overcome these challenges and to improve the efficacy of ChatGPT, we have uncovered some potential future directions.

Improving Conversational Capabilities, Personalization, Multimodal design, and Trustworthiness are some of the key areas that hold significant promise for the future of ChatGPT. By focusing on these directions and exploring possible solutions to current challenges, ChatGPT can become more ubiquitous and effective in the future, with the potential to revolutionize how humans interact with technology and can become a more effective and trustworthy tool for various segments of society.

We posit that this review will inspire further research and development to fully harness the vast potential of ChatGPT across diverse application areas for both the researchers working in this field and the users exploring the use of ChatGPT. Undoubtedly, it is just the beginning of the ChatGPT-era and we anticipate that future GPT will bring significant benefits to our lives in the near future. We look forward to seeing the continued evolution and growth of ChatGPT research in the years to come.

\end{document}